\documentclass[runningheads]{llncs}

\usepackage{eccv}

\usepackage{eccvabbrv}

\usepackage{graphicx}
\usepackage{booktabs}
\usepackage{multirow}
\usepackage{makecell}
\usepackage{capt-of}
\usepackage{bm}
\usepackage{hyphenat}
\usepackage[accsupp]{axessibility}  % Improves PDF readability for those with disabilities.

\usepackage{hyperref}

\usepackage{orcidlink}

\begin{document}

\title{A Plug-and-Play 2D Motion Interface for Real-World Motion Language Models} 

\titlerunning{A Plug-and-Play 2D Motion Interface for MoLMs}

\author{Kaname Yokoyama\inst{1} \and
Norimichi Ukita\inst{1}}
\newcommand{\meanstd}[2]{%
  \makecell[c]{#1\\[-0.4mm]{\scriptsize $\pm$#2}}%
}% TODO FINAL: Replace with an abbreviated list of authors.
\authorrunning{K. Yokoyama and N. Ukita}

\institute{Toyota Technological Institute, 
Nagoya, Aichi, Japan \\
\email{\{sd25444,ukita\}@toyota-ti.ac.jp}
}

\maketitle

\begin{abstract}
    Motion Language Models (MoLMs) typically understand human motions by tokenizing 3D motion and processing the resulting tokens using a language model. However, obtaining accurate 3D motions from monocular videos is challenging, limiting their real-world applicability. To address this issue, we introduce a plug-and-play 2D Motion Interface that enables 3D-pretrained MoLMs to accept 2D motion inputs without modifying or fine-tuning the original models.
    Experiments on public datasets show that our method achieves performance comparable to 3D motion inputs across multiple MoLMs and outperforms training MoLMs from scratch on 2D motions. We further construct a monocular real-world video motion evaluation dataset and introduce a real-video adapter, demonstrating the usefulness of 2D motions over 3D motions under the evaluated monocular pose-estimation setting. These results suggest that 2D motion provides a practical interface for deploying MoLMs in real-world motion understanding settings. Code is available at \url{https://github.com/irajisamurai/2D-Motion-Interface}.
  %Motion Language Model（MoLM）**はよく、3次元（3D）の人体動作をトークン化し、そのトークン列を言語モデルで処理することで人間の動作を理解する。しかし、単眼動画から高精度な3D動作を取得することは困難であり、これが実世界での適用性を制限している。 そこで，本研究では3D動作で学習されたMoLMsに対し、元のモデルの変更やファインチューニングを行うことなく2D動作を入力可能にするプラグアンドプレイ型の2D Motion Interfaceを導入する．
  %公開データセットを用いた実験では、提案手法は複数のMoLMにおいて3D動作入力と同等の性能を達成し、さらに2D動作のみを用いてMoLMをゼロから学習する方法を上回る性能を示した。また、単眼の実世界動画からなる動作評価データセットを新たに構築し、実動画用アダプタを導入した。その結果、評価対象とした単眼姿勢推定の設定において、3D動作よりも2D動作を用いる方が有用であることを示した。これらの結果は、\textbf{2D動作が、実世界における動作理解タスクへMoLMを適用するための、実用的なインターフェースとなり得ることを示唆している。コードおよびデータセットは補足資料に含まれている。

\end{abstract}

\section{Introduction}
\label{sec:intro}

\begin{figure}[t]
    \centering
    \includegraphics[width=100mm]{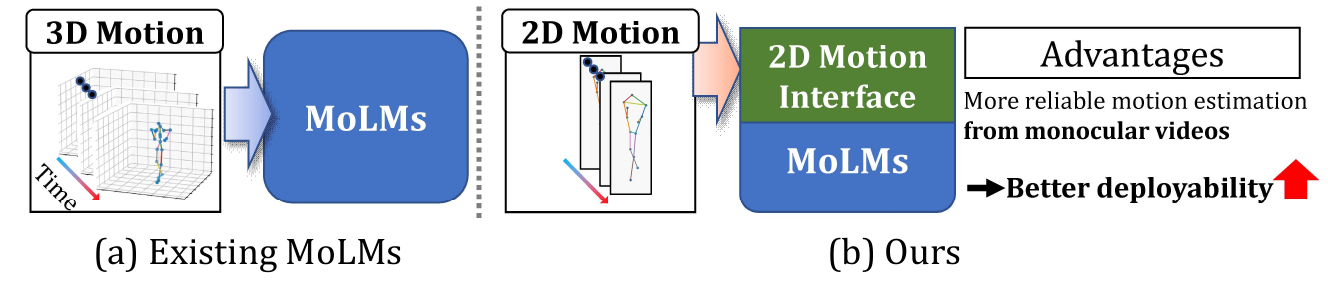}
    \caption{While existing MoLMs typically take 3D motions as input, the proposed 2D Motion Interface (i.e., 2D motion encoder) enables existing MoLMs based on VQ-VAE motion tokenization to accept 2D motion inputs. Since 2D pose estimation is generally easier than 3D pose estimation, our approach improves the deployability of MoLMs in real-world applications.
    }
    \label{fig:abstract}
    \vspace{-5mm}
\end{figure}
    %既存のMoLMsは一般的に3D動を入力とする一方，2D motion interface(2D motion encoder)はVQ-VAEによるモーションの離散トークン化を用いる既存のMoLMsに対して，2D motionの入力を可能にします．一般的に，2D姿勢推定は3D姿勢推定よりも容易であるため，我々の提案手法はMoLMsの実応用性を向上させます．

Human motion understanding has advanced rapidly alongside the development of Large Language Models (LLMs)~\cite{llmsaregoodaction,mgpt2,mgpt3,mlm_daily,chatpose,t2mgpt,t2mgpt_hifi,mgpt4}. In particular, Motion Language Models (MoLMs), such as MotionGPT~\cite{mgpt}, which learn to discretize motion from 3D motion datasets~\cite{humanml3d, aist,kit}, have achieved strong performance in tasks such as motion captioning and motion generation (Fig.~\ref{fig:abstract}(a)). Since these models are trained on 3D motions reconstructed from motion capture systems or multi-view observations, it is commonly assumed that their motion-related capabilities are grounded in 3D motion. However, this assumption has not been explicitly verified.

%人間動作理解は大規模言語モデル(LLMs)とともに急速に発展している．具体的には，3D動作データセットからモーションの離散トークン化を学習するMotionGPTのようなモーション言語モデル(MoLMs)はモーションキャプショニングやモーション生成において強力なパフォーマンスを達成している．これらはモーションキャプチャーシステムや多視点映像からの再構成された3D動作によって学習されているので，一般的には，これらのモデルの動作関連の能力は3D動作に基づいていると考えられる．しかしながら，この仮説は明示的に検証されているわけではない．

This assumption poses a significant challenge for real-world applications that leverage the motion understanding capabilities of MoLMs, such as action recognition and human-centered video understanding. In practical scenarios, human motions are typically estimated from videos captured by a monocular camera and then fed into MoLMs. However, the 3D motions commonly used in MoLMs~\ref{tab:humanml3d} include not only the relative 3D coordinates of each joint with respect to the root joint, but also joint rotation and global positions. Therefore, estimating equivalent 3D motions from monocular videos both accurately and efficiently remains challenging. The gap from the high-quality 3D motions used during training can degrade motion understanding performance.

%%この仮説は行動認識，人間中心の動画理解のようなMoLMsの動作理解能力を用いた実世界応用においては大きな課題となる．なぜなら，実世界応用では単眼カメラから得られた映像から動作を推定してMoLMsに入力すると考えられるが，MoLMsで一般的に用いられる3D動作には単に各関節のrootからの相対3D座標だけでなく各関節の回転角やグローバル座標を含まれる．そのため，単眼映像から同等の3D動作を高速，かつ高精度に推定することは難しく，学習時の高品質な3D動作とのギャップが動作理解性能を低下させる要因となる．

Therefore, this paper challenges this assumption. If the motion representation required in MoLMs can be sufficiently approximated for motion understanding using 2D motions derived from 2D poses (e.g., the COCO keypoints~\cite{coco}), 3D-pretrained MoLMs could perform motion understanding even from monocular 2D observations. This would substantially improve the practicality and real-world applicability of MoLMs.
%そこで，本論文ではこの仮定に疑問を投げかける．もしCOCOキーポイント形式のような 2D姿勢からなる2D動作からでもMoLMsで用いられる動作表現を動作理解のために十分近似可能であれば，単眼カメラによる2D観測のみであっても，3D動作で事前学習されたMoLMsにおいて動作理解が実現でき，MoLMsの実応用性を格段に向上させることができる．

Moreover, examining this assumption also offers insight into the extent to which motion understanding relies on 3D-specific information such as 3D joint rotation and depth. We note that our analysis observes this under a setting where 2D features are explicitly aligned to the 3D latent space, rather than establishing that 3D-specific information is unnecessary in general.
%さらに，この仮定を検証することは，joint rotationやdepthといった3D固有の情報に，動作理解がどの程度依存しているのかについての知見を与える．なお，本分析は，2D特徴を3D潜在空間に明示的にアラインする設定においてこれを観察するものであり，3D固有の情報が一般に不要であることを示すものではない．
To examine this assumption, we introduce a plug-and-play 2D Motion Interface for MoLMs. The proposed interface is designed as an external module that can be attached to pretrained MoLMs without requiring any modification or fine-tuning of the base MoMLs. Specifically, it is implemented as a 2D motion encoder that projects 2D motions into the continuous latent space of the discrete tokenizer (VQ-VAE~\cite{vq-vae}) used by MoLMs (Fig.~\ref{fig:abstract}(b)).

%この疑問に答えるために，本論文ではMoLMs向けのplug-and-Play 2D Motion Interfaceを提案する．このインタフェースは，事前学習済みのMoLMsに対して改変やファインチューニングを一切必要としないモジュールとして設計されている。具体的には， 2D動作をMoLMsの離散トークナイザー(VQ-VAE)の 連続潜在空間へ射影する2D動作エンコーダとして実装される．

With this 2D motion encoder, MoLMs can directly accept 2D motions as input. Furthermore, because the encoder shares the same latent space as the discrete tokenizer, it enables direct comparisons between the latent motion features and motion tokens derived from 3D motions and those derived from 2D motions. This unified representation space allows us to experimentally investigate whether the motion representations used in MoLMs can be effectively approximated from 2D motions. 
To further validate the practical advantages of 2D motions over 3D motions, we construct a real-world video dataset in which each video is paired with its corresponding estimated 2D motions, estimated 3D motions, and motion captions. Since video-derived motions contain distribution shifts caused by camera-parameter-dependent perspective projection and pose-estimation noise, we further propose a real-video adapter to bridge this gap and improve the robustness of MoLMs in real-world applications.
%この2D動作エンコーダによってMoLMsは2D動作を入力として受け入れることが可能になる．また，2D動作エンコーダは離散トークナイザーと潜在空間を共有しているため3D動作と2D動作から得られる潜在動作特徴および動作トークンを直接比較することが可能となり，2D動作を用いてMoLMsで用いられる動作表現を近似可能かどうかを実験的に検証できる．
%2D動作を用いることの実用的な優位性をさらに検証するため，本研究では，各動画に対応する推定2D動作，推定3D動作，およびモーションキャプションをペアとして含む実動画データセットを構築する．実動画から推定された動作には，依然としてカメラパラメータに依存する透視投影によるドメインシフト，推定ノイズが含まれるため，本研究ではこのギャップを埋めるための実動画アダプターをさらに提案する．このアダプターにより，実世界応用におけるMoLMsの実用性とロバスト性を向上させる．

Our contributions are summarized as follows:
\begin{enumerate}

    \item We introduce a plug-and-play 2D Motion Interface (i.e., a 2D motion encoder) that enables 3D-pretrained MoLMs to accept 2D motion inputs. The proposed interface requires neither modification nor fine-tuning of the pretrained MoLMs and can be implemented with a very low training cost.
    
    \item Through experiments on multiple public datasets and MoLM baselines, we demonstrate that 2D motion inputs achieve performance comparable to that of 3D motion inputs. Furthermore, we show that 2D and 3D motions exhibit strong consistency in both the continuous latent space and the discrete token space, indicating that 2D motions can effectively approximate the motion representations used by 3D-pretrained MoLMs for motion understanding.
    
    \item We constructed a real-world video dataset and proposed a real-video adapter. Under the evaluated monocular pose-estimation setting, experiments using estimated 3D and 2D motions as inputs show that 2D motions achieve higher motion understanding performance with lower computational cost than 3D motion estimation. These results demonstrate the practical advantages of using 2D motions in real-world applications.

\end{enumerate}

%我々の貢献は次のように要約される
%１．3D動作で事前学習されたMoLMsに2D動作を入力可能にする非常にシンプルなplug-and-Play 2D Motion Interface(2D動作エンコーダ）を提案します．MoLMsの改変やファインチューニングは必要なく，非常に低い学習コストで実装することができます．
%2.  モーションキャプショニングタスクにおいて複数のベースライン，データセットにおいて2D動作エンコーダを用いることで3D動作を入力として用いた場合と同等の性能を2D動作入力で達成した．また，2D動作と3D動作は連続潜在空間および離散トークン空間において動作理解において十分な整合性が得られることを示した．
%3.  実動画データセットを構築し，加えて実動画アダプターを提案する．推定された3D，2D動作を入力として用いた場合，2D動作の方が3D動作推定よりも低い計算コストで動作理解性能が高いことを示し，実応用において2D動作を用いることの優位性を示した．

\section{Related Work}
\label{sec:related}

\subsection{Motion Language Models}
TM2T~\cite{TM2T} demonstrates that 3D human motions can be represented as language-like discrete token sequences through motion tokenization, making tasks such as motion generation and motion captioning feasible within an autoregressive framework. Subsequently, MotionGPT~\cite{mgpt} and MG-MotionLLM~\cite{mg-motionllm} incorporate pretrained language models and instruction tuning, enabling motions and text to be processed in a unified language modeling paradigm. MoLMs have also been extended in various directions.

For example, MoMask~\cite{momask} adopts masked generative modeling over discrete motion tokens, improving controllability and generation quality.  By tokenizing additional modalities (e.g., music/audio) and feeding all modality tokens into a pretrained LLM, M$^{3}$ GPT~\cite{m3gpt} unifies multiple modalities and tasks in a single model. RapVerse~\cite{rapverse} jointly generates vocals and whole-body motion by combining language, audio, and motion tokens. Furthermore, SocialGen~\cite{socialgen} and several subsequent studies~\cite{intergen,TSMM} have extended motion language modeling to multi-person social interactions, enabling the generation and understanding of collaborative behaviors involving multiple individuals.

Despite the remarkable progress of MoLMs, most existing studies rely on 3D motion datasets such as HumanML3D, which contain rich 3D information including global positions and joint rotations.  In contrast, the observations available in real-world applications are often limited to monocular videos, from which estimating equivalent 3D motions remains challenging. This gap between the training data and real-world observations constitutes a major obstacle to the practical deployment of MoLMs for motion understanding tasks.

%TM2Tは3D動作をVQ-VAEによって言語のような離散トークン列に変換することで，モーション生成やモーションキャプショニングタスクを実行可能であることを示した．MotionGPTやMG-MotionLLMにおいては事前学習済み言語モデルやInstruction-tuningを導入し，言語モデルにおいて動作とテキストを統一的に扱うことを可能にしている．
%また，研究は複数の方向に発展しており，モーショントークン列に対する学習目標の改良，追加モダリティへの拡張，さらには複数人物の動作や相互作用を含むより複雑な状況への対応が進められている．

%例えば，MoMaskは離散動作トークンに対するマスク生成モデリングを採用し，生成品質および制御性を向上させている。 
%また，音楽や音声といった追加モダリティをトークン化し，すべてのモダリティのトークンを事前学習済みLLMへ入力することで，M\$^\{3\}$GPTは複数モダリティと複数タスクを単一モデルで統合している。一方，RapVerseは言語・音声・動作トークンを組み合わせることで，歌声と全身動作を同時に生成する。
%さらに，SocialGenや複数の研究ではモーション言語モデリングを複数人物の社会的相互作用へ拡張し，複数人による協調動作の生成および理解を可能にしている。

%このようにMoLMsは大きく発展してきたものの，既存研究のほとんどはHumanML3Dのような高精度な3D動作データを前提としている．これらのデータは各関節の相対3D座標に加えてグローバル座標や関節回転情報を含む．一方，実世界で利用可能な観測情報は単眼カメラ映像であることが多く，そこから同等の3D動作を推定することは困難である．この学習データと実環境とのギャップが，動作理解におけるMoLMsの実応用性を制限する大きな要因となっている． 

\subsection{2D/3D Motion Representation and Cross-Modal Alignment}
Bridging 2D observations and 3D motion is a long-standing goal in the community. One direction is reconstructing 3D motion from monocular or multi-view inputs~\cite{3dpose_estim1,3dpose_estim2,3dpose_estim3,3dpose_estim4,3dpose_estim5}. Recent work suggests that 3D motion cues can emerge even without explicit 3D supervision. Free3D~\cite{free3d} shows that 3D motion can be learned from monocular 2D supervision using 3D-free regularizations, such as view consistency and physical plausibility.
Another line of work studies representations shared across 2D and 3D. In~\cite{unified2d3d}, a unified codebook shared between 2D and 3D pose spaces is proposed. V-VIPE~\cite{v-vipe} learns a canonical, view-invariant embedding for 3D pose, enabling both 2D and 3D inputs to be encoded into a shared space. More closely related to our setting, \cite{distill2d3d} decouples and distills 3D latent features to improve the robustness of discriminative 2D skeleton action recognition.

Inspired by these works, we explore whether MoLMs trained on 3D motions can be adapted to 2D inputs for real-world motion understanding. Specifically, our encoder aligns 2D motions to the latent space of a \emph{frozen} MoLMs, enabling motion-language understanding from 2D inputs without any fine-tuning of the base MoLMs.
%
%2D観測と3D人体動作のギャップを埋めることは，長年にわたり研究コミュニティにおける重要な課題であった。その代表的なアプローチの一つが，単眼あるいは多視点入力から3D動作を復元する研究である。
%近年では，明示的な3D教師信号を用いなくても3D動作情報を獲得できる可能性が示されている。例えば，Free3Dは，視点間整合性や物理的妥当性といった3D不要の正則化を利用することで，単眼2D教師信号のみから3D人体動作を学習できることを示した。
%また，2Dと3Dの間で共有される表現に着目した研究も存在する。文献では，2D姿勢空間と3D姿勢空間の両方で共有される統一コードブックが提案されている。さらに，V-VIPEは，3D姿勢に対する視点不変な正準埋め込み表現を学習することで，2D入力と3D入力を共通の特徴空間へ写像することを可能にしている。

%これらの研究にインスパイアされて，我々は実世界での動作理解ために3D動作で学習されたMoLMsが2D入力に適用可能かを調査します．具体的には～

\section{Proposed Method}
\label{sec:method}

To enable MoLMs to accept 2D motion inputs, we introduce a simple \textbf{2D motion encoder} as a plug-and-play \textbf{2D Motion Interface}. With the proposed encoder, existing MoLMs can directly perform motion understanding tasks, such as motion captioning, from 2D motions.
 %MoLMsに2D動作を入力可能にするために，2D Motion Interfaceとしてシンプルな2D motion encoderを提案します． 2D motion encoderによって既存のMoLMsの動作理解タスク(例：モーションキャプショニング）を2D動作から実行することを可能にします．

\begin{figure}[t]
    \centering
    \includegraphics[width=\linewidth]{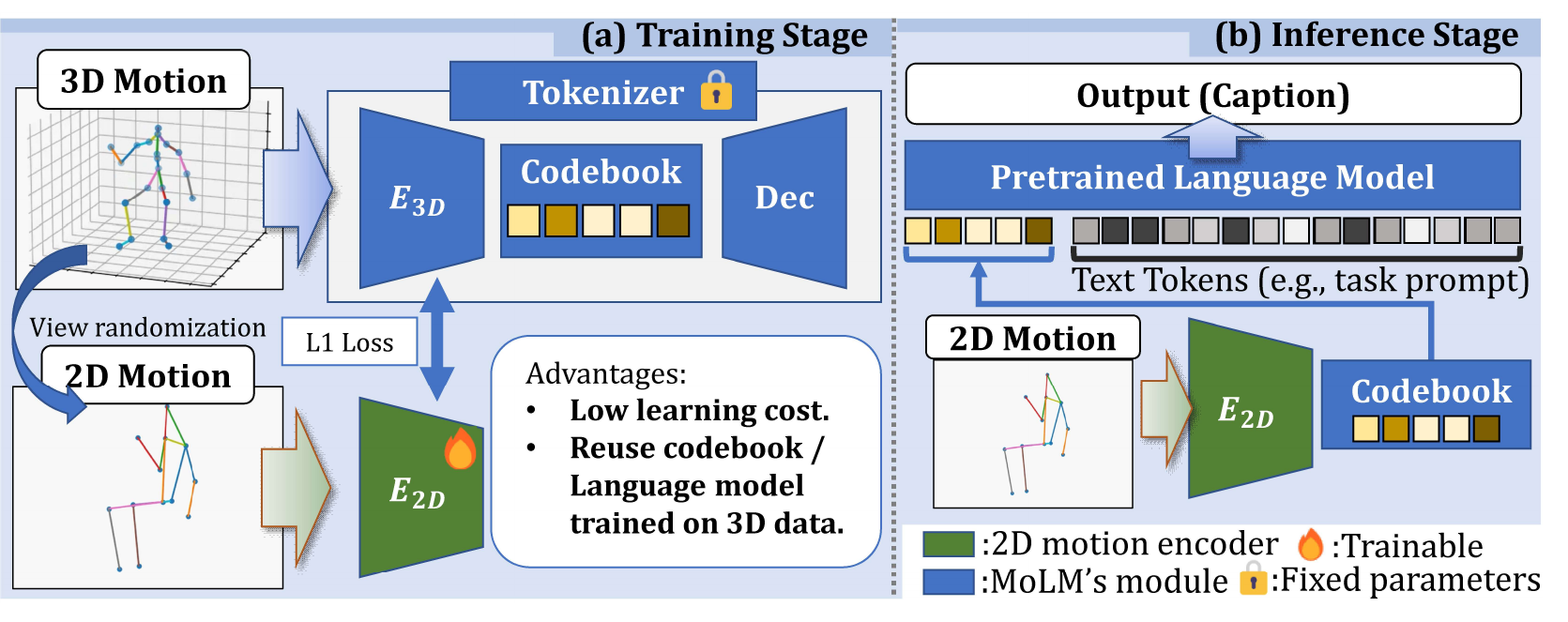}
     \caption{\textbf{Overview of the proposed pipeline.} During training, the pretrained MoLM remains frozen and only the 2D motion encoder is trained. Specifically, the encoder is trained using an L1 feature-matching loss to align 2D motion features with the continuous latent space of the VQ-VAE encoder. During inference, the resulting 2D motion features are quantized into discrete motion tokens using the VQ-VAE codebook and fed into the pretrained language model.
    %提案手法のパイプライン．学習時にはMoLMsのモジュールは固定し，2D motion encoderを学習する．具体的にはL1特徴マッチング損失を用いて，MoLMsのVQ-VAEのエンコーダのEが用いる連続潜在空間へ 整列する．
    %推論時には2D動作をMoLMsのVQ-VAEのコードブックによって離散トークン化し，事前学習済み言語モデルへ入力する．
    }
    \label{fig:pipeline}
\end{figure}

Figure.~\ref{fig:pipeline} illustrates the training and inference pipeline of the proposed 2D motion encoder. As shown in Fig.~\ref{fig:pipeline}(a), the backbone MoLM is pretrained on 3D motion data (Sec.~\ref{subsec:preliminries}) and remains frozen throughout training. Only the 2D motion encoder, shown in the lower branch, is optimized. Specifically, the encoder is trained by minimizing an alignment loss between the 3D motion features produced by the VQ-VAE encoder of the MoLM and the 2D motion features produced by the 2D motion encoder (Sec.~\ref{subsec:2dencoder}).  Through this alignment, the 2D motion encoder learns to map 2D motions (Sec.~\ref{subsec:2dinputs}) into the continuous latent space of the MoLM, allowing existing MoLMs to operate directly on 2D motion inputs without any architectural modifications or fine-tuning.
During inference (Fig.~\ref{fig:pipeline}(b)), the resulting 2D motion features are quantized into discrete motion tokens using the VQ-VAE codebook and subsequently fed into the pretrained language model to perform motion understanding tasks (Sec.~\ref{subsec:infrence}).

%2D motion encoderの学習，推論パイプラインを図？に示します．図?(a)に示すようにベースとなるMoLMsは3D動作によって事前に学習されており，学習時にはそのパラメータを固定します．一方，下段に示す2D motion encoderのみを学習します． 具体的には，MoLMsのVQ-VAEのエンコーダの出力である3D動作特徴と2D motion encoderのを出力でアライメント損失を最小化するように学習することで，2D motion encoderが2D動作をMoLMsのVQ-VAEのエンコーダの連続潜在空間への写像を可能にします．
%推論時には2D motion encoderから得られた2D動作特徴をMoLMsのVQ-VAEのコードブックによって離散トークン化し，MoLMsの事前学習済み言語モデルに入力することで動作理解タスクを実行します．

\subsection{Preliminaries: MoLMs}
\label{subsec:preliminries}
\paragraph{Overview of Target MoLMs.}
In this work, we focus on MoLMs that represent human motions as \textbf{discrete motion token sequences}, analogous to words in natural language. These models typically consist of a VQ-VAE that serves as a motion tokenizer and a language model (e.g., T5~\cite{t5}) that processes the resulting motion token sequences. 

%本研究では，人間の動作をテキストにおける単語に対応する離散的な動作トークン列として表現するMoLMsを対象とする．これらのモデルは一般に，動作のtokenizerとして機能するVQ-VAEと，動作トークン列を処理する言語モデル（例：T5）から構成される． 

\paragraph{3D motion representation.}
In MoLMs~\cite{mg-motionllm,mgpt}, the human body at each frame is represented as a 263-dimensional state vector. This state vector consists of the following components: 
(i) root rotation velocity (1D),
(ii) root linear velocity (2D),
(iii) root $y$ position (1D),
(iv) joint positions ($21\times 3=63$D),
(v) joint rotations ($21\times 6=126$D),
(vi) joint velocities ($22\times 3=66$D), and
(vii) foot contact (4D).
A motion sequence of $T$ frames is denoted as $\bm{X}^{\text{3D}}\in\mathbb{R}^{T \times 263}$.

%ほとんどのMoLMsでは各フレームにおける人体の状態を，263次元の状態ベクトルとして表現する。この状態ベクトルは，以下の情報から構成される。

%内訳（省略）

%長さ T フレームの動作系列を，X3D∈R263×T\textbackslash{}bm\{X\}^\{\text\{3D\}} \textbackslash{}in \textbackslash{}mathbb\{R\}^\{263 \textbackslash{}times T\}X3D∈R263×T と表す。 

\paragraph{VQ-VAE tokenization.}
The VQ-VAE consists of an encoder $E_{\mathrm{3D}}$, a decoder $D_{\mathrm{3D}}$, and a codebook $B _{\mathrm{3D}}=\{\bm{b}_k\}_{k=1}^{K}$, where $K$ denotes the size of the codebook.

When a 3D motion $\bm{X}^{3D}$ is input to the encoder $E_{\mathrm{3D}}$, a latent feature sequence $\bm{H}^{\mathrm{3D}}=[\bm{H}^{\mathrm{3D}}_{1}, \ldots, \bm{H}^{\mathrm{3D}}_{S}] = E_{\mathrm{3D}}(\bm{X}^{\mathrm{3D}}) \in \mathbb{R}^{S \times D}$ is obtained, where $S$ is the temporal length after downsampling by the encoder and $D$ is the latent feature dimension.
Next, each latent feature vector is quantized by assigning it to the nearest codebook entry, resulting in a discrete token sequence $\bm{C}^{\mathrm{3D}}=[c_1,c_2,\ldots,c_S]$. Here, each token $c_i$ denotes the index of the assigned codebook entry.
The VQ-VAE is trained using a reconstruction loss together with the standard VQ-VAE embedding loss and commitment loss~\cite{vq-vae,mgpt}. 

%VQ-VAEはエンコーダ$E_\text{3D}$，デコーダ$D_\text{3D}$，コードブック$B_\text{3D} = {b_k}_{k=1}^K$から構成されます．ここでKはコードブックのサイズです．
%3D動作$\bm{X}^{\text{3D}}$がエンコーダ$E_\text{3D}$に入力されると，潜在特徴$\bm{H}^{\text{3D}} = [\bm{H}^{\text{3D}}_{1}, \dots, \bm{H}^{\text{3D}}_{S}] = E_\text{3D}(\bm{X}^{\text{3D}}) \in \mathbb{R}^{D \times S}$が得られます．ここで，SSS はエンコーダによるダウンサンプリング後の時系列長，DDD は潜在特徴の次元数を表す。 
%次に，各潜在特徴ベクトルをコードブック中の最も近いコードへ割り当て，離散トークン列 C3D=[c1,c2...cs] を生成する。ここでcはコードブックのインデックスを表す．
%このVQ-VAEは，標準的なVQ-VAEの埋め込み損失およびコミットメント損失とともに，復元損失を用いて学習される。

\paragraph{How motion tokens are used by language models.}
The discrete token sequence $\bm{C}^{\mathrm{3D}}$ is processed by a language model to perform motion understanding tasks such as motion captioning. Many MoLMs provide a unified framework for motion understanding and generation through bidirectional translation between motion and text, namely motion-to-text and text-to-motion. Since the proposed 2D motion encoder is designed to interface 2D motions with existing MoLMs, we focus on the motion-to-text setting throughout this work.

%離散トークン列 \$\bm\{C\}^\{\mathrm\{3D\}}\$ は言語モデルに入力され，モーションキャプショニングなどの動作理解タスクが実行される．多くのMoLMsはmotion-to-textとtext-to-motionの双方向変換を通じて，動作理解と動作生成を統一的に扱う．一方，本研究で提案する2D motion encoderは2D動作を入力として既存のMoLMsへ接続することを目的としているため，本研究ではmotion-to-textタスクに焦点を当てる．  

 \subsection{2D motion with View Randomization}
\label{subsec:2dinputs}
The 2D motion sequence is derived from the 3D motion $\bm{X}^{3D}$ defined in Sec.~\ref{subsec:preliminries}. Specifically, 3D-specific cues, including depth and joint rotation information, are removed while preserving only the information observable from a monocular camera, resulting in a 2D motion  $\bm{X}^{2D}$. The construction procedure is described as follows:

\begin{enumerate}
    \item \textbf{Skeleton extraction.} From $\bm{X}^{3D}$, we reconstruct a 3D joint skeleton $\bm{J}^{\mathrm{3D}}\in\mathbb{R}^{T\times22\times3}$ in the world coordinate system. The reconstructed skeleton follows the 22-joint SMPL-style skeleton~\cite{Loper2015SMPL} used in HumanML3D~\cite{humanml3d}.
    \item \textbf{View randomization.} To encourage viewpoint-robust encoding, we randomly rotate $\bm{J}^{\text{3D}}$ by yaw (rotation about the vertical axis) and pitch (rotation about the horizontal axis) during training.
    \item \textbf{2D projection.} The rotated skeleton is projected onto a 2D plane by discarding the depth coordinate. We then retain only the 13 joints shared by both the SMPL and COCO~\cite{coco} keypoint formats, resulting in a 2D joint skeleton $\bm{J}^{\mathrm{2D}} \in \mathbb{R}^{T\times 13 \times 2}$.  To mimic the output of standard monocular pose estimators, we adopt the COCO keypoint format. We use orthographic projection to avoid camera-parameter assumptions, and handle the gap to perspective real-video observations with the real-video adapter in Sec.~\ref{subsec:noise_adapter}.
    \item \textbf{Per-frame 2D features.} From $\bm{J}^{\text{2D}}$, we construct a HumanML3D-like per-frame feature vector composed of:
    (i) root linear velocity (2D),
    (ii) root $y$ position (1D),
    (iii) joint positions ($13\times 2=26$D, root-relative),
    (iv) joint orientations (13D, root-relative), and
    (v) joint velocities ($13\times 2=26$D),
    which totals 68 dimensions. A motion sequence of $T$ frames is denoted as $\bm{X}^{\text{2D}}\in\mathbb{R}^{T \times68}$. The root joint is defined as the midpoint of the left and right hips, and all features are normalized.

\end{enumerate}
%2D動作は3D動作から構築される．具体的にはSec. \~\ref\{subsec:preliminries\} で定義した3D動作$\bm{X}^{\text{3D}}$から，深度情報や3D回転情報といった2D姿勢推定からは得られない情報を除去しつつ2D動作$\bm{X}^{\text{2D}}$を構築する．構築方法は以下のとおりである．

%1.スケルトンの抽出．3D動作X3Dから，世界座標における3D関節スケルトンを復元する．ここで用いるスケルトンは，HumanML3D\~\cite\{humanml3d\}で使用されている22関節のSMPL形式スケルトン\~\cite\{Loper2015SMPL\}である。
%2.視点変化に対して頑健な特徴表現を獲得するため，学習時には J3D\textbackslash{}bm\{J\}^\{\text\{3D\}}J3D にランダムな回転を適用する。具体的には，鉛直軸周りの回転（yaw）および水平軸周りの回転（pitch）をランダムに付与する。
%3.回転後のスケルトンから深度方向の座標を除去することで2D平面へ正投影する。その後，SMPL形式とCOCO形式\~\cite\{coco\}の両方に共通して存在する13関節のみを抽出し，J2D∈R13×2×T\textbackslash{}bm\{J\}^\{\text\{2D\}} \textbackslash{}in \textbackslash{}mathbb\{R\}^\{13 \textbackslash{}times 2 \textbackslash{}times T\}J2D∈R13×2×Tを構築する。一般的な単眼姿勢推定器の出力を模擬するため，本研究ではCOCO形式のキーポイント表現を採用する．なお，一般的な単眼姿勢推定器の出力を模擬するため，本研究ではCOCO形式のキーポイント表現を採用する．
%4.フレームごとの2D動作特徴．（省略）

  \subsection{2D motion encoder}
\label{subsec:2dencoder}
To map a 2D motion  $\bm{X}^{2D}$ into the continuous latent space of the VQ-VAE used by the MoLM, we introduce a 2D motion encoder $E_\text{2D}$. When a 2D motion  $\bm{X}^{2D}$ is input to the encoder $E_{\mathrm{2D}}$, a latent feature sequence $\bm{H}^{\mathrm{2D}}=[\bm{H}^{\mathrm{2D}}_{1}, \ldots, \bm{H}^{\mathrm{2D}}_{S}] = E_{\mathrm{2D}}(\bm{X}^{\mathrm{2D}}) \in \mathbb{R}^{S \times D}$ is obtained.
The encoder is trained by minimizing an L1 feature-alignment loss between the latent features generated by the VQ-VAE encoder $E_{\mathrm{3D}}$ and those produced by the 2D motion encoder $E_\text{2D}$.

\begin{equation}
\mathcal{L}_{\text{align}}
=
\frac{1}{S \, D}
\left\lVert \bm{H}^{\text{2D}} - \bm{H}^{\text{3D}} \right\rVert_{1}
\label{eq:objective}
\end{equation}
\vspace{-7mm}

%2D動作$\bm{X}^{\text{2D}}$をMoLMsのVQ-VAEのエンコーダの連続潜在空間へ写像するために2D motion encoderを導入します．（2D motion encoderの式省略）

%2D motion encoderはVQ-VAEのエンコーダE3Dの出力と2D motion encoderのl1損失を最小化することによって学習されます．（損失の式省略）

  \subsection{Inference}
\label{subsec:infrence}
During inference, a 2D motion  $\bm{X}^{2D}$ is first mapped to latent features $\bm{H}^{\mathrm{2D}}$ by the proposed 2D motion encoder $E_\text{2D}$. The latent features are then quantized into a discrete token sequence $\bm{C}^{\mathrm{2D}}$ using the VQ-VAE codebook $B_{\mathrm{3D}}$. The resulting token sequence $\bm{C}^{\mathrm{2D}}$ is subsequently fed into the language model $LM$ of an existing MoLM, either alone or together with a task prompt $\bm{P}$, to obtain the generated textual output $\hat{\bm{Y}} = LM\!\left(\bm{P}, \bm{C}^{\mathrm{2D}}\right)$. 

%推論時には，まず2D動作を2D motion enccoderに入力して潜在特徴H2Dを得る．次にMoLMsのVQ-VAEのコードブックB3Dを用いて量子化する．各潜在特徴ベクトルをコードブック中の最も近いコードへ割り当て，離散トークン列 C2D=[c1,c2...cs] を生成する．その後，生成された離散トークン列を単体，もしくはプロンプトとともにMoLMsの言語モデルへ入力し生成されたテキストを得る．
 
\section{Experiments}
\label{sec:experiments}
In this section, we evaluate the proposed 2D motion encoder on multiple motion understanding tasks and datasets. Section. ~\ref{subsec:exp_setup} introduces the datasets, evaluation metrics, and implementation details. Section. ~\ref{subsec:results} compares the proposed method against 3D motion inputs and MoLMs trained from scratch on 2D motions. Finally, Sec.~\ref{subsec:analysis} presents a detailed analysis of the proposed method. Qualitative results are provided in the Appendix.
%このセクションでは，2D motion encoderを複数のモーション理解タスク，データセットで評価します．データセット，評価指標，実装の詳細についてはSec.4.1 ．Sec. 4.2では3D動作での結果，2DからMoLMs全体をスクラッチで学習した場合の結果，2D motion encoderを用いた場合の結果を比較します．最後に2D motion encoderの分析をSec.4.3で実施します．定性結果はAppendixで提供されます．

\subsection{Experimental Setup}
\label{subsec:exp_setup}

\noindent\textbf{Datasets.} Following~\cite{mg-motionllm}, we conduct experiments on two motion-language datasets. The first is HumanML3D~\cite{humanml3d}, one of the largest motion-language datasets currently available. HumanML3D is built upon AMASS~\cite{amass} and consists of 14,616 motion sequences paired with 44,970 sequence-level text captions. This dataset is used for the motion captioning task. The second dataset is FineMotion~\cite{finemotion}, which re-annotates the captions in HumanML3D with more fine-grained descriptions. Specifically, each motion sequence is divided into snippets at fixed temporal intervals, resulting in 420,968 motion snippets. Each snippet is paired with a detailed caption describing the movements of body parts. This dataset is used for the motion-to-detailed-text task.

%実験は~\ciee{mg-motionllm}に従って，2つのモーション言語データセットで実施されます．一つ目は最も大きなモーション言語データセットであるHumanML3Dです．HumanML3Dは，AMASS\~\cite\{amass\}を基に構築されたデータセットであり，14,616個の動作系列と，それらに対応する44,970個の系列レベルのテキストキャプションから構成されます．このデータセットはモーションキャプショニングに使われます．2つ目はFineMotionです．FineMotionはHumanML3Dのキャプションをより詳細なキャプションにre-labelしたものです．具体的には、各モーションは一定の時間間隔でスニペットに分割され、420968個のスニペットが生成されます。各スニペットには、身体部位の動きに関する詳細な説明がペアになっています.このデータセットはmotion-to-detailed-textタスクに使われます．

\noindent\textbf{Metrics.} 
Following prior work~\cite{mg-motionllm,mgpt}, we evaluate motion captioning using R-Precision (Top-1/2/3), a retrieval-based metric for measuring motion-text matching accuracy, MM-Dist for assessing motion-text alignment, and standard language generation metrics, including BLEU~\cite{bleu}, ROUGE-L~\cite{rouge}, CIDEr~\cite{cider}, and BERTScore~\cite{bertscore}. For the motion-to-detailed-text task, we follow the evaluation protocol of~\cite{mg-motionllm} and report the aforementioned language generation metrics at both the sequence and snippet levels. 

%先行研究に従い，モーションキャプショニングでは，動作とテキストの対応付け精度を測定する検索ベース評価指標R-Precision（Top-1/2/3），動作と生成テキストのアライメントを評価するMM-Dist，およびBLEU~\cite{bleu}，ROUGE-L~\cite{rouge}，CIDEr~\cite{cider}，BERTScore~\cite{bertscore}といった標準的な言語生成評価指標を用いて評価する．motion-to-detailed-textタスクでは，\cite{mg-motionllm}に従い前述の言語生成評価指標を用いて評価する．評価は動作系列全体に対する詳細キャプションおよびスニペットレベルのキャプションの両方について実施する．

 \noindent\textbf{Implementation Details.}
The latent feature dimension of the 2D motion encoder is set to 512, and the temporal downsampling rate is set to 4, following prior works~\cite{mgpt,mg-motionllm}. The resulting latent sequence length is therefore $S=T/4$. The 2D motion encoder is trained using the AdamW optimizer with a learning rate of $1\times10^{-4}$, $\beta=(0.9, 0.99)$, and a batch size of 64. No learning rate scheduler is employed, and the model is trained for 3000 epochs. All experiments are conducted on a single NVIDIA A100 40GB GPU. For fair comparison, all reported results are obtained by running the corresponding models in our environment. For the proposed method, we report the mean and standard deviation over three runs in Tabs.~\ref{tab:humanml3d} and~\ref{tab:finemotion}. Additional details are provided in the Appendix.
%MoLMsのVQ-VAEのアーキテクチャは\cite{motiongpt}に統一し，それに合わせて，2D motion encoderの潜在特徴次元は512次元，時間方向のダウンサンプリング率を4とする．エンコード後の系列長は S=T/4S=T/4S=T/4 となる。 
%2D motion encoderはAdamW optimizerを用いて学習される．学習率は学習率を 1×10−41\textbackslash{}times10\^{-4\}1×10−4，β=(0.9,0.99)\textbackslash{}beta=(0.9,0.99)β=(0.9,0.99)，バッチサイズを64，weight decayを0.0に設定する。また，学習率スケジューラは使用せず，3000エポック学習を行う。 学習は1枚のNVIDIA A100-40GB GPU上で実施された．公正な比較のため，すべての結果は，我々の環境でモデルを実行した結果得られたものである．追加の詳細はAppendixで提供されます．

\subsection{Results of Motion understanding Tasks}
\label{subsec:results}
\begin{table}[t]
\centering
\caption{Comparison of motion captioning results on HumanML3D test set. The best and second-best results within each model are highlighted in bold and underlined, respectively. Avg Drop vs 3D denotes the average percentage difference from the 3D Input across all metrics. The MotionGPT 2D Scratch result is omitted due to reproducibility issues; see the Appendix for details.}
\label{tab:humanml3d}
\small
\resizebox{\linewidth}{!}{%
\begin{tabular}{llccc ccccccc}
\toprule
\multirow{2}{*}{Model} & \multirow{2}{*}{Method}
& \multicolumn{3}{c}{R-Prec$\uparrow$}
& MM-Dist$\downarrow$ & BLEU-1$\uparrow$ & BLEU-4$\uparrow$ & ROUGE-L$\uparrow$ & CIDEr$\uparrow$ & BERTScore$\uparrow$ & \makecell{Avg Drop\\vs 3D$\uparrow$}\\
\cmidrule(lr){3-5}
& & Top-1 & Top-2 & Top-3 & & & & & & &\\
\midrule
\multirow{3}{*}{TM2T}
& 3D Input   
& \textbf{0.488} & \textbf{0.689} & \underline{0.786} 
& \underline{3.137} 
& \underline{0.615} & \textbf{0.233} & \textbf{0.492} & \textbf{0.691} & \textbf{0.368} & -\\
& 2D Scratch 
& 0.279 & 0.447 & 0.562 
& 4.921 
& 0.550 & 0.175 & \underline{0.437} & 0.499 & \underline{0.312} & $-25.8\%$\\
& \textbf{Ours}       
& \meanstd{\underline{0.485}}{0.002} & \meanstd{\underline{0.683}}{0.003} & \meanstd{\textbf{0.788}}{0.004}
& \meanstd{\textbf{3.129}}{0.005} 
& \meanstd{\textbf{0.616}}{0.001} & \meanstd{\underline{0.232}}{0.000} & \meanstd{\textbf{0.492}}{0.000} & \meanstd{\underline{0.688}}{0.004} & \meanstd{\textbf{0.368}}{0.001} & \textbf{$-0.2\%$}\\
\midrule
\multirow{2}{*}{MotionGPT}
& 3D Input   
& \underline{0.516} & \underline{0.707} & \textbf{0.803} 
& \underline{2.994} 
& \underline{0.429} & \underline{0.062} & \underline{0.345} & \textbf{0.079} & \underline{0.316} & -\\
& \textbf{Ours}       
& \meanstd{\textbf{0.523}}{0.002} & \meanstd{\textbf{0.717}}{0.003} & \meanstd{\underline{0.802}}{0.003}
& \meanstd{\textbf{2.985}}{0.015} 
& \meanstd{\textbf{0.431}}{0.000} & \meanstd{\textbf{0.063}}{0.001} & \meanstd{\textbf{0.346}}{0.001} & \meanstd{\underline{0.077}}{0.001} & \meanstd{\textbf{0.319}}{0.001} & \textbf{$+0.4\%$}\\
\midrule
\multirow{3}{*}{MG-MotionLLM}
& 3D Input   
& \textbf{0.583} & \textbf{0.787} & \textbf{0.873} 
& \textbf{2.571} 
& \textbf{0.507} & \textbf{0.091} & \textbf{0.402} & \textbf{0.095} & \textbf{0.386} & -\\
& 2D Scratch 
& 0.487 & 0.667 & 0.758 
& 3.289 
& 0.494 & 0.078 & 0.385 & 0.074 & 0.364 & $-12.8\%$\\
& \textbf{Ours}       
& \meanstd{\underline{0.576}}{0.007} & \meanstd{\underline{0.770}}{0.001} & \meanstd{\underline{0.855}}{0.001}
& \meanstd{\underline{2.701}}{0.021} 
& \meanstd{\underline{0.499}}{0.001} & \meanstd{\underline{0.087}}{0.001} & \meanstd{\underline{0.396}}{0.001} & \meanstd{\underline{0.090}}{0.001} & \meanstd{\underline{0.379}}{0.001} & \textbf{$-2.8\%$}\\
\bottomrule
\end{tabular}}
\vspace{-3mm}
\end{table}

\noindent\textbf{Comparison on motion captioning.}
Motion captioning is a task that generates a textual description corresponding to a given motion sequence. Table.~\ref{tab:humanml3d} reports the results on TM2T, MotionGPT, and MG-MotionLLM. For each model, we compare three settings: using 3D motion inputs, training the entire MoLM from scratch on 2D motions, and using the proposed 2D motion encoder. Additional details are provided in the Appendix.
Overall, the proposed method achieves performance comparable to 3D motion inputs across most text-motion alignment metrics, including R-Precision and MM-Dist, as well as language generation metrics, despite relying only on monocular 2D motion inputs. In addition, compared with training MoLMs from scratch on 2D motions, our method substantially reduces the performance gap from the 3D-input setting. These results demonstrate that the proposed 2D motion encoder is superior to straightforwardly training MoLMs from scratch on 2D motions in terms of both performance and computational cost, confirming the effectiveness of our approach.
%モーションキャプショニングは，与えられたモーション系列に対応する言語的描写を生成するタスクである．Tab.\~\ref\{humanml3d\}では，TM2T，MotionGPT，MG-MotionLLMを対象に，3D動作を入力した場合，MoLM全体を2D動作からスクラッチで学習した場合，および提案する2D motion encoderを用いた場合の結果を比較する．追加の詳細はAppendixに示す．

%全体として，提案手法は単眼2D動作のみを入力とするにもかかわらず，R-PrecisionやMM-Distといったtext-motion alignment指標，および言語生成評価指標の多くにおいて，3D動作入力と同等の性能を達成している．さらに，MoLM全体を2D動作からスクラッチで学習した場合と比較して，3D動作入力からの性能低下を大幅に抑えている．これはstraight forwardにMoLMsを2D動作からスクラッチで学習するよりも，性能面でも計算コストの面でも2D motion encoderが優位であることを示しており提案手法の有効性を確認できる． 

\begin{table}[t]
\centering
\caption{Comparison of motion-to-detailed-text results on the FineMotion test set.}
\label{tab:finemotion}
\small
\resizebox{95mm}{!}{%
\begin{tabular}{ll cccccc}
\toprule
Level & Method
& BLEU-1$\uparrow$ & BLEU-4$\uparrow$ & BLEU-7$\uparrow$ & ROUGE-L$\uparrow$ & BERTScore$\uparrow$ & \makecell{Avg Drop\\vs 3D$\uparrow$}\\
\midrule
\multirow{3}{*}{Sequence}
& 3D Input   & \textbf{0.828} & \textbf{0.667} & \textbf{0.536} & \textbf{0.651} & \textbf{0.523} & -\\
& 2D Scratch & \underline{0.814} & 0.637 & 0.497 & 0.629 & 0.483 & $-4.9\%$\\
& \textbf{Ours} & \meanstd{\textbf{0.828}}{0.001} & \meanstd{\underline{0.665}}{0.001} & \meanstd{\underline{0.533}}{0.001} & \meanstd{\underline{0.648}}{0.001} & \meanstd{\underline{0.517}}{0.001} & \textbf{$-0.5\%$}\\
\midrule
\multirow{3}{*}{Snippet}
& 3D Input   & \textbf{0.672} & \textbf{0.478} & \textbf{0.354} & \textbf{0.606} & \textbf{0.509} & -\\
& 2D Scratch & 0.653 & 0.456 & 0.329 & 0.588 & 0.487 & $-4.4\%$\\
& \textbf{Ours} &\meanstd{\underline{0.666}}{0.000} & \meanstd{\underline{0.473}}{0.001} & \meanstd{\underline{0.349}}{0.001} & \meanstd{\underline{0.601}}{0.001} & \meanstd{\underline{0.502}}{0.001} & \textbf{$-1.1\%$}\\
\bottomrule
\end{tabular}}
\vspace{-7mm}
\end{table} 

\noindent\textbf{Comparison on motion-to-detailed-text.}
Motion-to-detailed-text is a task that generates more detailed textual descriptions for a given motion sequence than standard motion captioning. Following the evaluation setting of~\cite{mg-motionllm}, Tab.~\ref{tab:finemotion} reports the results using MG-MotionLLM as the base model. Similar to the results on motion captioning, the proposed method achieves performance comparable to 3D motion inputs on many language generation metrics at both the sequence and snippet levels. These results indicate that the proposed 2D motion encoder can effectively preserve not only the coarse semantics of the entire motion sequence but also fine-grained motion information required for more detailed motion descriptions.

 %motion-to-detailed-textは，与えられたモーション系列に対して，通常のmotion captioningよりも詳細な言語的描写を生成するタスクである．Tab.\~\ref\{humanml3d\}では，MG-MotionLLMをベースモデルとして用いた結果を報告する．Motion captioningにおける結果と同様に，シーケンスレベルとスニペットレベルの両方において提案手法は多くの言語生成評価指標において，3D動作入力と同等の性能を達成している．この結果は，提案する2D motion encoderが，モーション全体の大まかな意味だけでなく，より詳細な動作記述に必要な細粒度な動作情報も有効に保持できることを示している．

\subsection{Analysis of 2D motion encoder}
\label{subsec:analysis}

\begin{figure}[t]
\centering
\begin{minipage}[t]{0.30\linewidth}
\vspace{0pt}
\centering
\includegraphics[width=\linewidth]{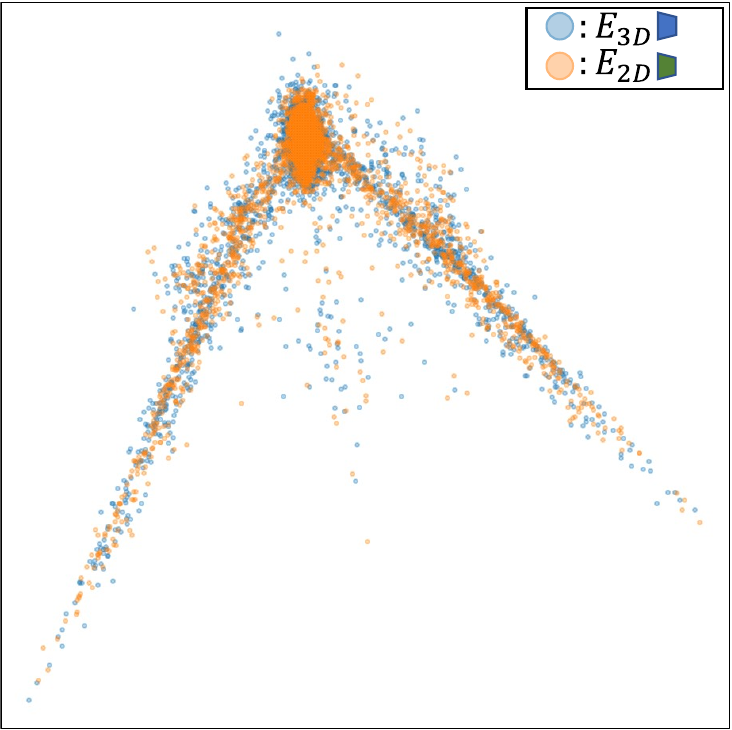}
\caption{PCA visualization of latent features obtained from 3D and 2D motions.}
\label{fig:pca_motiongpt}
\end{minipage}
\hfill
\begin{minipage}[t]{0.68\linewidth}
\vspace{0pt}
\centering
\captionof{table}{Robustness to neighboring token replacement on motion captioning performance.}
\label{tab:token_replace}
\scriptsize
\setlength{\tabcolsep}{3pt}
\resizebox{\linewidth}{!}{%
\begin{tabular}{lcccc}
\toprule
\multirow{2}{*}{Method} & R-Prec$\uparrow$ & \multirow{2}{*}{BLEU-1$\uparrow$} & \multirow{2}{*}{BERTScore$\uparrow$} & \multirow{2}{*}{\makecell{Avg Drop\\vs 3D$\uparrow$}} \\
\cmidrule(lr){2-2}
& Top-1 & & & \\
\midrule
3D Input &\textbf{0.516} &\underline{0.429}  &0.316  &-  \\
k=1 &0.502 & 0.428& 0.317& -0.9\%\\
k=3 &0.515 & \underline{0.429}& \textbf{0.320}& +0.4\%\\
k=5 & 0.511& \textbf{0.430}& \underline{0.318}& 0.0\%\\
k=7 & 0.510& \underline{0.429}& 0.313& -0.7\%\\
k=9 & 0.500& 0.426& 0.315& -1.4\%\\
random & 0.223& 0.357& 0.224& -34.2\%\\
\bottomrule
\end{tabular}%
}
\end{minipage}
\end{figure}
\noindent\textbf{Spatial Alignment between 2D and 3D Motion Distributions.}
In Sec.~\ref{subsec:results}, we showed that the proposed 2D motion encoder can preserve the motion understanding performance of MoLMs even when their 3D motion inputs are replaced with 2D motion inputs. To understand why this is possible, this section analyzes the latent features and discrete token sequences obtained from 2D and 3D motions. 

Specifically, given a 3D motion $\bm{X}^{3D}$ and its corresponding 2D motion  $\bm{X}^{2D}$, we first feed $\bm{X}^{3D}$ into the VQ-VAE encoder $E_{\mathrm{3D}}$ to obtain the 3D latent features $\bm{H}^{\mathrm{3D}}$, and feed  $\bm{X}^{2D}$ into the 2D motion encoder to obtain the 2D latent features $\bm{H}^{\mathrm{2D}}$. We then quantize each latent feature using the VQ-VAE codebook $B_{3D}$, yielding the discrete token sequences $\bm{C}^{\mathrm{3D}}$ and $\bm{C}^{\mathrm{2D}}$. For the latent feature analysis, we apply PCA to $\bm{H}^{\mathrm{3D}}$ and project both $\bm{H}^{\mathrm{3D}}$ and $\bm{H}^{\mathrm{2D}}$ into the resulting low-dimensional space to visualize the overlap between their distributions. Since quantitative distances between latent features are difficult to interpret directly, we additionally evaluate Top-$k$ token agreement in the discrete token space. Through this analysis, we examine how well the 2D motion encoder reproduces the continuous latent space and discrete token space of MoLMs. The experiments are conducted using the VQ-VAE of MotionGPT and its corresponding 2D motion encoder. 

%Sec.~\ref{subsec:results}では，MoLMsへの入力を3D動作から2D動作へ置き換えた場合でも，提案する2D motion encoderによって動作理解性能を維持できることを示した．本節では，その理由を明らかにするため，2D動作および3D動作から得られる潜在特徴と離散トークン列を分析する．
%具体的には，3D動作 $\bm{X}^{3D}$ と，それに対応する2D動作  $\bm{X}^{2D}$ が与えられたとする．まず， $\bm{X}^{3D}$ をVQ-VAEエンコーダ $E_{\mathrm{3D}}$ に入力して3D潜在特徴 $\bm{H}^{\mathrm{3D}}$ を取得し， $\bm{X}^{2D}$ を2D motion encoderに入力して2D潜在特徴 $\bm{H}^{\mathrm{2D}}$ を取得する．次に，VQ-VAEのコードブック $B_{3D}$ を用いてそれぞれの潜在特徴を量子化し，離散トークン列 $\bm{C}^{\mathrm{3D}}$ および $\bm{C}^{\mathrm{2D}}$ を得る．
%潜在特徴の分析では，$\bm{H}^{\mathrm{3D}}$ に対してPCAを適用して得られる低次元空間へ $\bm{H}^{\mathrm{3D}}$ と $\bm{H}^{\mathrm{2D}}$ を射影し，両者の分布の重なりを可視化する．また，離散トークン列の分析では，$\bm{C}^{\mathrm{3D}}$ と $\bm{C}^{\mathrm{2D}}$ の一致率をTop-1およびTop-3 accuracyとして報告する．これにより，2D motion encoderが3D MoLMsの連続潜在空間および離散トークン空間をどの程度再現できているかを検証する．実験はMotionGPTのVQ-VAEとそれに対応する2D motion encoderを用いて実施する． 

As shown in Fig.~\ref{fig:pca_motiongpt}, the distributions of the latent features $\bm{H}^{\mathrm{3D}}$ obtained from 3D motions and $\bm{H}^{\mathrm{2D}}$ obtained from 2D motions largely overlap. This result qualitatively indicates that the 2D motion encoder $E_{\mathrm{2D}}$ can map 2D motions into the continuous latent space formed by the VQ-VAE encoder $E_{\mathrm{3D}}$. Furthermore, the agreement between the discrete token sequences reaches 44.8\% for Top-1, 59.6\% for Top-2, and 67.1\% for Top-3. Considering that the codebook size is $K=512$, the expected agreement rate when randomly selecting tokens is only $1/K \simeq 0.20\%$ for Top-1 and $3/K \simeq 0.59\%$ for Top-3. These results indicate that the proposed 2D motion encoder successfully aligns 2D motions with the continuous latent space of MoLMs, and that this alignment is reflected in the quantized token space as high agreement with 3D-derived motion tokens.

%Fig.~\ref{fig:pca_motiongpt}に示すように，3D動作から得られる潜在特徴 $\bm{H}^{\mathrm{3D}}$ と，2D動作から得られる潜在特徴 $\bm{H}^{\mathrm{2D}}$ の分布は大きく重なっている．この結果は，2D motion encoder $E_{\mathrm{2D}}$ が，2D動作をVQ-VAEエンコーダ $E_{\mathrm{3D}}$ によって形成される連続潜在空間へ写像できていることを定性的に示している．さらに，離散トークン列の一致率はTop-1で44.8\%，Top-2で59.6\%，Top-3で67.1\%であった．コードブックサイズが $K=512$ であることを考えると，ランダムにトークンを選択した場合の期待一致率はTop-1で $1/K \simeq 0.20\%$，Top-3で $3/K \simeq 0.59\%$ に過ぎない．以上の結果から，提案する2D Motion Encoderは，2D動作を3D MoLMsの連続潜在空間へ適切に整列できており，この整列が量子化後の離散トークン列においても3D動作由来のトークン列との高い一致率として反映されていることが確認できる． 

\noindent\textbf{Robustness to Neighboring Token Replacement.}
We showed that the Top-1 agreement between the discrete token sequences obtained from 2D motions and those obtained from 3D motions is 44.8\%, which is substantially higher than the random baseline. On the other hand, although the Top-1 agreement is below 50\%, Sec.~\ref{subsec:results} shows that high motion understanding performance is maintained when using 2D motion inputs. This result suggests that, even when exactly the same tokens are not selected, neighboring tokens in the VQ-VAE codebook may have similar semantics, causing only limited semantic degradation as inputs to the language model. To verify this hypothesis, this section conducts a token replacement experiment on the discrete token sequence $\bm{C}^{\mathrm{3D}}$ obtained from a 3D motion $\bm{X}^{\mathrm{3D}}$. Specifically, we randomly select 50\% of the tokens in the sequence and replace each selected token with a token randomly chosen from its $k$ nearest neighbors in the VQ-VAE latent space. The resulting partially replaced token sequence is then fed into the language model of the MoLM, and the change in motion captioning performance is evaluated. %Sec.~\ref{}では，2D動作から得られた離散トークン列と3D動作から得られた離散トークン列のTop-1一致率が44.8%であり，ランダムベースラインと比較して十分に高いことを示した．一方で，Top-1一致率が50%未満であるにもかかわらず，Sec.~\ref{}では2D動作入力によって高いmotion understanding性能が維持されている．この結果は，完全に同一のトークンが選択されなくても，VQ-VAEのコードブック内で近傍に位置するトークンが類似した意味を持ち，言語モデルへの入力としては意味的な破壊が小さい可能性を示唆している．本節では，この仮説を検証するため，3D動作 $\bm{X}^{\mathrm{3D}}$ から得られる離散トークン列 $\bm{C}^{\mathrm{3D}}$ に対して，トークン置換実験を行う．具体的には，離散トークン列からランダムに選ばれた50%のトークンについて，VQ-VAEの潜在空間における$k$近傍トークンの中からランダムに選択したトークンへ置き換える．こうして得られた置換後の離散トークン列をMoLMの言語モデルへ入力し，motion captioning性能の変化を評価する.

Table.~\ref{tab:token_replace} shows the results when $k$ is set from 1 to 9, as well as the result when tokens are randomly replaced without using k-nearest neighbors. When $k=1$--$9$, the performance degradation is very small. In contrast, when tokens are randomly replaced, Avg Drop vs 3D reaches $-34.2\%$, indicating a sharp performance decrease. This supports the hypothesis discussed above that neighboring tokens in the VQ-VAE codebook have similar semantics. Moreover, even when $k=9$, Avg Drop vs 3D is only $-1.4\%$, indicating high robustness to neighboring-token replacement. 
These results suggest that the high motion understanding performance can be maintained even when the Top-1 agreement between the discrete token sequences obtained from 2D and 3D motions is 44.8\%. This is likely because the VQ-VAE codebook contains multiple tokens with similar semantics, and as long as neighboring tokens are selected, the semantic degradation as input to the language model remains limited. 
%Tab.~\ref{}にk=1~9に設定した場合，また，kNNではなくランダムにトークンを入れ替えた場合の結果を示す．k=1~9の場合においては性能の低下は非常に軽微である．一方，ランダムにトークンを入れ替えた場合にはAvg Drop vs 3Dは-34.2\%であり，性能は急激に低下する．このことから，先ほどの述べたVQ-VAEのコードブック内で近傍に位置するトークンは類似した意味を持つという仮説が指示される．また，k=9においても，Avg Drop vs 3Dは-1.4%であるからロバスト性も高いことがわかる．これらの結果から，2D動作から得られた離散トークン列と3D動作から得られた離散トークン列のTop-1一致率が44.8%でも，高い動作理解性能が維持される要因としてVQ-VAEのコードブック内には類似した意味をもつトークンが複数存在するため，近傍のトークンさえ得られれば言語モデルへの入力としては意味的な破壊が小さい考えることができる．

\section{Evaluation on Monocular Real Videos}
\label{sec:realvideo}
In this section, we evaluate the motion understanding performance of MoLMs on monocular real videos to demonstrate the advantages of 2D motions over 3D motions in practical applications. We also propose a real-video adapter and show that it further improves the practicality of MoLMs. Section.~\ref{subsec:real_world_video} describes the construction of the real-world video dataset, Sec.~\ref{subsec:noise_adapter} introduces the real-video adapter, and Sec.~\ref{subsec:eval_real_video} compares inference results using 3D and 2D motions extracted from real-world videos.
%このセクションでは実応用における3D動作に対する2D動作の優位性を示すために実動画データセットにおけるMoLMsの動作理解性能を評価します．また，ノイズ軽減アダプターを提案し，それによってさらにMoLMsの実用性を高められることを示す．実動画データセットの構築方法についてはSec. 5.1， Sec. 5.2ではノイズ軽減アダプターの実装について，Sec. 5.3では実動画から抽出した3D，2D動作に対して推論を実施し，結果を比較します．

  \subsection{Monocular real-world video motion dataset}
\label{subsec:real_world_video}

In the experiments in Sec.~\ref{sec:experiments}, we used clean 3D motions and clean 2D motions obtained by projecting them into 2D. However, in practical applications, motions are estimated from monocular real-world videos and inevitably contain estimation noise. Therefore, to evaluate how 2D and 3D motion inputs perform under such realistic conditions, we construct a \textbf{monocular real-world video motion dataset}. 
%Sec.~\ref{sec:experiments}の実験では，クリーンな3D動作と，それを2Dへ投影して得られるクリーンな2D動作を用いた．しかし，実応用では，動作は単眼の実動画から推定されるため，必然的に推定ノイズを含む．そこで，このような実環境に近い条件下で2D動作入力と3D動作入力がどのように機能するかを評価するため，本研究では\textbf{単眼実動画動作データセット}を構築する．

Specifically, we randomly extract motion clips from HumanML3D~\cite{humanml3d} test set and present each clip to participants. The participants imitate the presented motions, and the performed motions are recorded using a monocular RGB camera (iPhone SE 2nd generation). The dataset consists of 132 videos recorded by 10 participants (male and female) ranging in age from their 20s to 50s. In total, the dataset contains 14,878 frames, corresponding to 743.9 seconds of video. During recording, we vary the relative viewpoints between the participants and the camera to include diverse observation conditions. For evaluation, we use the original text captions from HumanML3D corresponding to the sampled motion clips. We then estimate 3D and 2D motions from each video using TRACE~\cite{TRACE} and WHAM~\cite{WHAM} for 3D pose estimation and ViTPose~\cite{vitpose} for 2D pose estimation. The SMPL parameters estimated by a 3D pose estimator are converted into the 263-D representation described in Sec.~\ref{subsec:preliminries}, following HumanML3D~\cite{humanml3d}. Human verification showed that 86.4\% of the video-caption pairs were semantically consistent; details are provided in the Appendix.
Another possible way to obtain estimated 3D motions is 2D-to-3D lifting. However, we do not use this approach because it only provides 3D joint \textbf{positions} and does not provide the joint rotations required by MoLMs, as described in Sec.~\ref{subsec:preliminries}. 

%具体的には，HumanML3D~\cite{humanml3d}からランダムに動作クリップを抽出し，各クリップを被験者に提示する．被験者は提示された動作を模倣し，その様子を単眼RGBカメラ（iPhone SE 第2世代）で撮影する．データセットは，20代から50代までの男女10名の被験者によって撮影された132本の動画から構成される。データセット全体では，14,878フレーム，743.9秒分の動画を含む．また，撮影時には被験者とカメラの相対的な視点を変化させることで，多様な観測条件を含むようにした． 評価には，サンプリング元の動作クリップに対応するHumanML3Dの元のテキストキャプションを用いる．その後，3D姿勢推定モデルであるTRACEおよび2D姿勢推定モデルであるViTPose\~\cite\{vitpose\}を用いて，各動画からそれぞれ3D動作および2D動作を推定する. TRACEによって推定されたSMPLパラメータは~\cite{humanml3d}に従って，Sec.~/ref{subsec}の263-D表現に変換される．推定された3D動作を得る方法として，2D→3D liftingも考えられるが，この方法では関節\textbf{位置}しか与えず、Sec.~\ref{subsec:preliminries}で述べたようなMoLMが要求する joint rotation を供給しない ため実施していない．
   \subsection{Noise reduction for real-world videos}
\label{subsec:noise_adapter}

\begin{figure}[t]
    \centering
    \includegraphics[width=100mm]{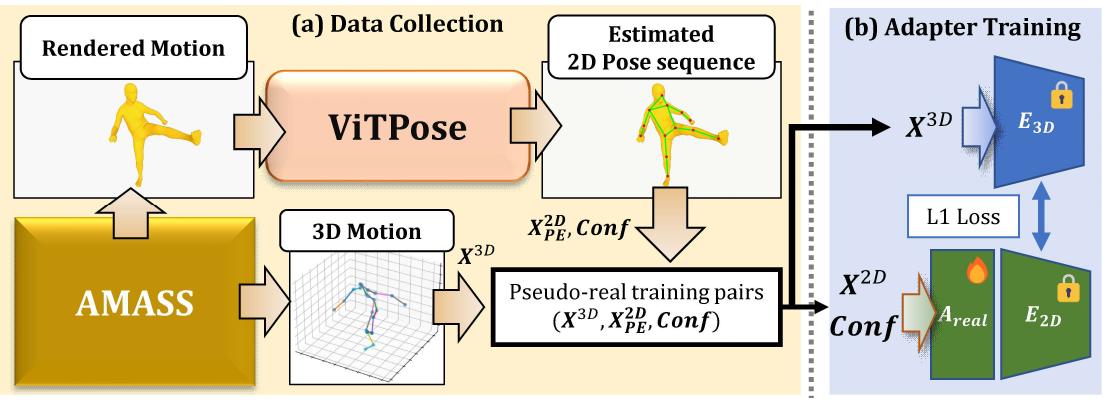}
     \caption{Training pipeline of the real-video adapter. We render 3D motions from random viewpoints and apply 2D pose estimation to the rendered videos to obtain estimated 2D motions $\bm{X}_{\mathrm{PE}}^{\mathrm{2D}}$ and joint confidence scores $\bm{Conf}$. During training, the adapter $A_{\mathrm{real}}$ is inserted before the 2D motion encoder, and only $A_{\mathrm{real}}$ is optimized using the same alignment loss as in Sec.~\ref{subsec:2dencoder}.
      %ノイズ軽減アダプターの学習パイプライン．3D動作をランダム視点からレンダリングし，2D姿勢推定を適用することで，推定2D動作と関節信頼度 \$\bm\{Conf\}$ を取得する．学習時には，2D motion encoderの前段にアダプター \$A\_{\textbackslash{}mathrm\{real\}}\$ を挿入し，Sec.\~\ref\{subsec:2dencoder\}と同様のアライメント損失を用いて \$A\_{\textbackslash{}mathrm\{real\}}\$ のみを最適化する．
      }
    \label{fig:noise_reduction}
    \vspace{-5mm}
\end{figure}

\noindent\textbf{Real-video adapter.}
To bridge the gap caused by the distribution mismatch between the orthographic-projection-based 2D motions used during training and the perspective-projection-based 2D observations obtained from real-world videos and pose-estimation noise, we propose a real-video adapter.
First, let $\bm{X}_{\mathrm{PE}}^{\mathrm{2D}} \in \mathbb{R}^{T \times 68}$ denote the 2D motion constructed from the COCO keypoint format 2D pose sequence $\bm{J}_{\mathrm{PE}}^{\mathrm{2D}}\in\mathbb{R}^{T \times 13\times 2}$ estimated from a monocular video by a pose estimator, following the same procedure as in Sec.~\ref{subsec:2dinputs}. We also denote the confidence scores of each joint by $\bm{Conf}\in\mathbb{R}^{T \times 13}$.

Next, to bring the estimated 2D motion $\bm{X}_{\mathrm{PE}}^{\mathrm{2D}}$ closer to the 2D motion distribution used during training, we introduce a real-video adapter $A_{\mathrm{real}}$ before the 2D motion encoder:
\begin{equation}
{\bm{H}'}^{\mathrm{2D}}
=
E_{\mathrm{2D}}(A_{\mathrm{real}}(\bm{X}_{\mathrm{PE}}^{\mathrm{2D}}, \bm{Conf}))
\in \mathbb{R}^{S \times D}.
\end{equation}

%動画から推定された2D動作に含まれる姿勢推定ノイズや，2D Motion Encoderの学習時に用いた正投影ベースの2D動作と実動画から得られる透視投影ベースの2D観測との間に生じる分布ギャップを軽減するため，本研究ではreal-video adapterを提案する．

%まず，X^2D_PEを姿勢推定器によって単眼動画から推定されたCOCO形式の2D姿勢系列J^2D_PEにSec. 3.2と同様の方法を用いて構築された2D motionとする．また，この時の各関節の信頼度をConfとする．
%次に姿勢推定によって得られた入力を学習時の2D入力分布へ近づけるための2D motion encoderの前にAdapterを導入する．

\noindent\textbf{Pseudo-real training pairs.}
A major challenge in training the adapter $A_{\mathrm{real}}$ is that real-world videos only provide pose-estimated 2D motions and are not annotated with corresponding ground-truth (GT) 3D motions. If pose-estimated 2D motions and their corresponding GT 3D motions were available, the adapter could be trained using the same objective function as that used for the 2D motion encoder, namely the alignment loss in Eq.~\ref{eq:objective}, to mitigate the noise.
%アダプタ ArealA\_\text\{real\}Areal を学習する上での大きな課題は，実動画から得られるのは姿勢推定された2D動作のみであり，対応する正解3D動作が付与されていないことである。
 %もし，姿勢推定によって得られた2D動作と，それに対応する正解3D動作が利用できれば，2Dエンコーダの学習に用いたものと同じ目的関数（すなわち，式(\ref{eq:objective}におけるアライメント損失 ） を用いてアダプタを学習することで， ノイズを軽減することができる．

To address this issue, we synthesize a training dataset consisting of GT 3D motions and their corresponding estimated 2D motions by leveraging 3D motions from clean 3D human motion datasets. In this work, we use the AMASS dataset as the source 3D human motion dataset.  Figure.~\ref{fig:noise_reduction}(a) illustrates the overview of the data collection process.
 Specifically, for each motion sequence in AMASS, we render videos from random viewpoints using Pyrender~\cite{pyrender}. We then apply a 2D pose estimator, ViTPose~\cite{vitpose}, to the rendered videos to obtain pose-estimated 2D skeletons in the COCO keypoint format, along with confidence scores for each joint.
 %そこで本研究では，クリーンな3D人体動作データセットに含まれる3D動作を利用して正解3D動作とそれに対応する2D動作からなる学習データセットを合成する．この3D人体動作データセットとして，本研究ではAMASSデータセットを用いる。図？に，データ収集プロセスの概要を示す。 

 %具体的には，AMASSの各動作系列に対して，Pyrender\~\cite\{pyrender\}を用いてランダム視点から動画をレンダリングする。その後，レンダリングされた動画に対して2D姿勢推定器（本研究ではViTPose\~\cite\{vitpose\}）を適用し，姿勢推定されたCOCO形式の2Dスケルトンおよび各関節の信頼度スコアを取得する。

 \noindent{\textbf{Training objective.}}
Figure.~\ref{fig:noise_reduction}(b) illustrates the training procedure of the adapter $A_{\mathrm{real}}$. During training, the 2D motion encoder $E_{\mathrm{2D}}$ is kept frozen, and only $A_{\mathrm{real}}$ is optimized. Following the objective in Eq.~\ref{eq:objective}, $A_{\mathrm{real}}$ is trained by minimizing the L1 feature-alignment loss between the latent features produced by $E_{\mathrm{3D}}$ and those produced by the 2D motion encoder.

\begin{equation}
\mathcal{L}_{\text{adapt}}
=
\frac{1}{S \, D}
\left\lVert {\bm{H}'}^{\mathrm{2D}} - \bm{H}^{\text{3D}} \right\rVert_{1}.
\label{eq:objective_adapter}
\end{equation}
Additional details on the architecture and training of the adapter are provided in the Appendix.

%図~\ref{fig:noise_reduction}にアダプタA_realの学習方法を示す．学習時には2D motion encoderE_2Dは固定し，A_realのみ学習する．A_realは式(\ref{ep:objective})と同様にE3Dの出力と2D motion encoderのl1損失を最小化することによって学習されます．

%adapterのアーキテクチャや学習についての追加の詳細はAppendixで提供されます．

%式は省略

\subsection{Evaluation results}
\label{subsec:eval_real_video}

\begin{table}[t]
\centering
\caption{Comparison of motion captioning results on monocular real-world video motion dataset.}
\label{tab:eval_realvideo}
\small
\resizebox{\linewidth}{!}{%
\begin{tabular}{llcccccc}
\toprule
Model & Method & BLEU-1$\uparrow$ & BLEU-4$\uparrow$ & ROUGE-L$\uparrow$ & CIDEr$\uparrow$ & BERTScore$\uparrow$ & \makecell{Avg Drop\\vs Ref 3D$\uparrow$}\\
\midrule
\multirow{6}{*}{TM2T}
& Ref 3D Input   
& 0.599 & 0.190 & 0.460 & 0.609 & 0.354 & -\\
\cmidrule(lr){2-8}
& 3D Input (TRACE)   
& 0.453 & 0.102 & 0.360 & 0.252 & 0.222 & $-37.7\%$\\
& 3D Input (WHAM)   
& 0.474 & 0.092 & 0.380 & 0.258 & 0.241 & $-35.9\%$\\
& 2D Input
& 0.554 & 0.147 & 0.419 & 0.403 & 0.285 & $-18.5\%$\\
& 3D Input (WHAM) w/ $A_{\mathrm{3D}}$        
& \underline{0.588} & \underline{0.166} & \underline{0.436} & \underline{0.486} & \underline{0.304} & $-10.8\%$\\
& 2D Input w/ $A_{\mathrm{real}}$       
& \textbf{0.619} & \textbf{0.218} & \textbf{0.461} & \textbf{0.645} &\textbf{0.344} & \textbf{$+4.3\%$}\\

\midrule
\multirow{6}{*}{MotionGPT}
& Ref 3D Input   
& 0.461 & 0.069 & 0.405 & 0.149 & 0.352 & -\\
\cmidrule(lr){2-8}

& 3D Input (TRACE) 
& 0.369 & 0.034 & 0.315 & 0.044 & 0.218 & $-40.3\%$\\
& 3D Input (WHAM) 
& 0.380 & 0.035 & 0.339 & 0.056 & 0.264 & $-34.1\%$\\
& 2D Input
& 0.361 & \textbf{0.047} & 0.321 & 0.075 & 0.236 &$-31.4\%$\\
& 3D Input (WHAM) w/ $A_{\mathrm{3D}}$        
& \underline{0.408} & \underline{0.046} & \underline{0.365} & \underline{0.102} & \underline{0.285} & $-21.1\%$\\
& 2D Input w/ $A_{\mathrm{real}}$       
& \textbf{0.435} & 0.045 & \textbf{0.367} & \textbf{0.105} & \textbf{0.306} & \textbf{$-18.5\%$}\\

\midrule
\multirow{6}{*}{MG-MotionLLM}
& Ref 3D Input   
& 0.556 & 0.117 & 0.450 & 0.233 & 0.436 & -\\
\cmidrule(lr){2-8}

& 3D Input (TRACE) 
& 0.385 & 0.033 & 0.310 & 0.058 & 0.240 & $-50.7\%$\\
& 3D Input (WHAM) 
& 0.433 & \underline{0.064} & 0.350 & 0.095 & 0.308 & $-35.6\%$\\
& 2D Input
& 0.392 & 0.044 & 0.321 & 0.062 & 0.264 & $-46.7\%$\\
& 3D Input (WHAM) w/ $A_{\mathrm{3D}}$        
& \underline{0.469} & 0.036 & \underline{0.378} & \underline{0.106} & \underline{0.352} & $-34.9\%$\\
& 2D Input w/ $A_{\mathrm{real}}$       
& \textbf{0.507} & \textbf{0.093} & \textbf{0.411} & \textbf{0.151} & \textbf{0.374} & \textbf{$-17.5\%$}\\

\bottomrule
\vspace{-10mm}
\end{tabular}}
\end{table}

For evaluation on the monocular real-world video motion dataset, we use the language generation metrics introduced in Sec.~\ref{subsec:exp_setup}. We compare estimated 3D motions, estimated 2D motions, and estimated 2D motions with the real-video adapter. For fairness, we also evaluate an adapter $A_{\mathrm{3D}}$ for the estimated 3D motions, using the same architecture and objective as the real-video adapter but changing only the input dimension to 263, without confidence inputs. 
As a reference, we report the original HumanML3D motions sampled for video collection as Ref 3D Input.  Note that Ref 3D Input is not the strict GT 3D motion corresponding to the motion performed in the real-world video. This allows us to evaluate the performance degradation caused by using 3D/2D motions estimated from real-world videos, as well as the effectiveness of the proposed real-video adapter.

 \begin{figure}[t]
    \centering
    \includegraphics[width=\linewidth]{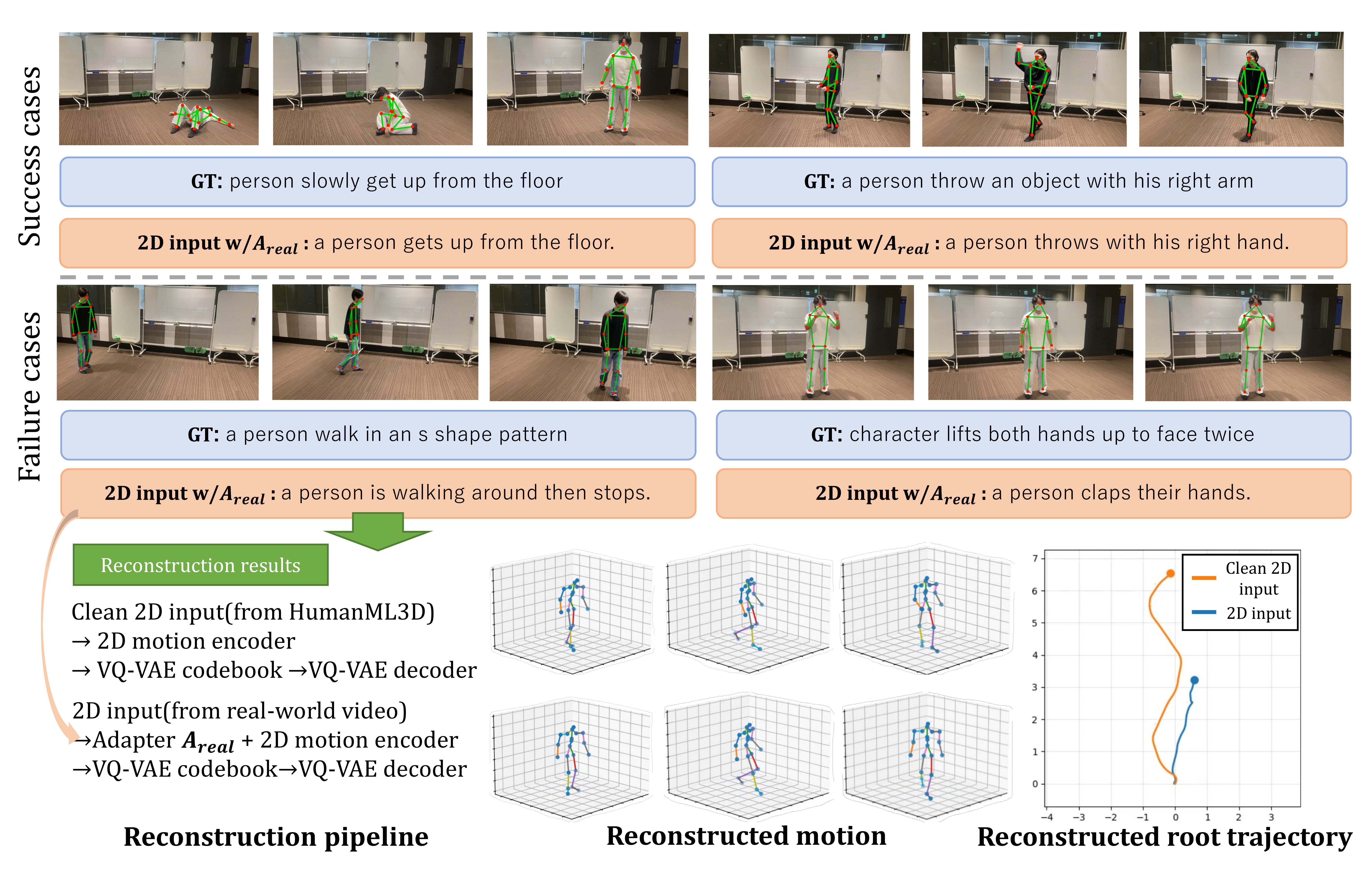}
     \caption{Qualitative results on monocular real-world video motion dataset. For reconstruction, we compare two inputs: the estimated 2D motion used as input (2D Input), and the clean 2D motion constructed from the corresponding Reference 3D motion following the same procedure as in Sec.~\ref{subsec:2dinputs} (Clean 2D Input).
     %再構成では，推定された2D動作を入力する場合（2D input)とその元となった3D動作からSec.~\ref{sec:method}と同様に2D動作を構築し，入力する場合(Clean 2D input)を比較する．
      }
    \label{fig:real_video}

\end{figure}

\noindent\textbf{Comparison on estimated 2D/3D inputs.}

Table.~\ref{tab:eval_realvideo} presents the quantitative results. Here, 3D Input w/ $A_{\mathrm{3D}}$ is applied to WHAM, which showed promising performance among the 3D Input settings without the adapter. First, we observe that 2D Input w/ $A_{\mathrm{real}}$ outperforms all 3D Input setting on almost all metrics across all models.
In addition, WHAM, a recent 3D pose estimation method, is substantially more computationally expensive than the combination of ViTPose and the adapter $A_{\mathrm{real}}$ used in our 2D Input setting. Specifically, excluding human detection, WHAM requires 250.2 GFLOPs per frame, whereas ViTPose (Base) requires only 17.2 GFLOPs.
These results support the advantage of using 2D motions over 3D motions in practical applications. Furthermore, 2D Input w/ $A_{\mathrm{real}}$ consistently outperforms 2D Input w/o $A_{\mathrm{real}}$ for all models, demonstrating the effectiveness of the proposed real-video adapter. 
%Tab.~\ref{tab:eval_realvideo}に定量評価の結果を示す．まず，全てのモデルにおいて全メトリクスで，2D Input w/ A_realが3D Inputの値を上回っていることが確認できる．また，最新の3D姿勢推定手法であるWHAMはそもそも，内部でViTPoseを用いているため，2D Inputで用いたViTPoseとアダプターA_realを合わせた場合と比較しても計算コストが大きい．具体的には，人検出を除いた場合，一フレーム当たりWHAMは250.2GFLOPs，一方，ViTPose (Base)は17.2 GFLOPsである．
%この結果は実応用における3D動作に対する2D動作の優位性を指示している．また，全てのモデルで2D Input w/o A_realが2D Input w/o A_realを上回っており，我々の提案する実動画アダプターの有効性を確認することができる．Fig.~\ref{}の上段にはMotionGPTの2D Input w/ A_realの定性結果を示している．全体としてGTキャプションと近いキャプションを出力できており，主要な動作については捉えられていることがわかる．

%Tab.~ref{}に定量的な結果を示す，3D Input w/ A_realはアダプターなしで有望な結果を残したWHAMに適用している．まず，全てのモデルにおいてほぼ全てのmtricsで2D Input w/ A_realが3D Inputを用いた場合の値を上回っていることが確認できる．In addition, WHAM, a recent 3D pose estimation method, is substantially more computationally expensive than the combination of ViTPose and the adapter $A_{\mathrm{real}}$ used in our 2D Input setting. Specifically, excluding human detection, WHAM requires 250.2 GFLOPs per frame, whereas ViTPose (Base) + $A_{\mathrm{real}}$ requires only 17.2 GFLOPs.
%この結果は実応用における3D動作に対する2D動作の優位性を支持している．また，全てのモデルで2D Input w/o A_realが2D Input w/o A_realを上回っており，我々の提案する実動画アダプターの有効性を確認することができる．Fig.~\ref{}の上段にはMotionGPTの2D Input w/ A_realの定性結果を示している．全体としてGTキャプションと近いキャプションを出力できており，主要な動作については捉えられていることがわかる．

\noindent\textbf{Limitation.}
When comparing Ref 3D Input and 2D Input w/ $A_{\mathrm{real}}$ in Tab.~\ref{tab:eval_realvideo}, MotionGPT and MG-MotionLLM show larger Avg Drop vs Ref 3D than TM2T. This may be because newer MoLMs can generate more detailed motion descriptions, while the correction process by the adapter smooths the motion sequence, suppressing fine-grained motion differences and making the representation closer to an average motion pattern.
The bottom part of Fig.~\ref{fig:real_video} shows failure cases. In the left example, fine-grained trajectory information, namely walking in an S-shaped path, is lost. In the right example, although the actual motion involves raising both hands near the face, the model incorrectly generates a more generic hand motion, i.e., clapping. To visualize information loss in the real-video inference pipeline, we feed the token sequences obtained from the 2D inputs into the VQ-VAE decoder and visualize the reconstructed motions and root trajectories. The reconstruction results show that, when the adapter is used, the reconstructed root trajectory no longer preserves the S-shaped path, indicating that such fine-grained information is lost. Another possible factor is imitation error during dataset construction, where discrepancies between the original Ref 3D motions and the actual performed motions may lead to semantic mismatches.
%一方で，Tab.~\ref{tab}においてRef 3D Inputと2D Input w/ $A_{\mathrm{real}}$ を比較すると, MGPTとMG-MOtionLLMはAvg Drop vs Ref 3DがTM2Tと比較して大きい．これは，新しいMoLMsほどより詳細な動作描写を生成できる一方で，アダプターによる補正の過程で動作系列を平滑化することで，細かな動作差分が失われ，平均的な動作表現に近づいてしまうためであると考えられる．

%Fig.~\ref{fig:real_video}の下段に，失敗例を示す．左の例では「S字に歩く」という細かな軌跡情報が失われ，右の例では，実際には両手を顔の近くへ上げる動作であるにもかかわらず，より一般的な手の動作であるclappingとして誤って生成されている．情報損失を可視化するため，2D入力から得られたトークン列をVQ-VAE decoderに入力し，再構成された動作とroot trajectoryを可視化している．実際に，アダプターを用いた場合には再構成されたroot trajectoryにはS字状の軌跡が確認できず，情報が失われていることがわかる．

%そのほかの原因としては，monocular real-world video motion dataset作成時に，被験者が提示されたクリップを完璧に模倣することは難しいので，GTの3D動作と差異が生まれ意味的に異なる動作として認識されることなどが考えられる．

\section{Conclusion}
We introduced a plug-and-play 2D Motion Interface that maps 2D motions into the continuous latent space of existing VQ-VAE-based MoLMs, enabling them to accept 2D motion inputs without modifying the original models. Experiments across multiple MoLMs showed performance comparable to 3D motion inputs. We further constructed a monocular real-world video motion dataset and introduced a real-video adapter, demonstrating that estimated 2D motions are more effective than estimated 3D motions under the evaluated monocular pose-estimation setting.

As a future research direction, from the perspective of practical deployment, it is important to address the over-generalization of generated captions, as discussed in Sec.~\ref{subsec:eval_real_video}. From the perspective of extending model capabilities, further training the language models of MoLMs on large-scale captioned human-motion video data is expected to expand the motion vocabulary and enable more fine-grained motion understanding.

%本論文ではMoLMsの実応用性を向上させるために既存のMoLMsに2D動作を入力可能にするPlug-and-Play 2D Motion Interface，具体的には2D動作をMoLMsのVQ-VAEのエンコーダの連続潜在空間へ写像するために2D motion encoderを提案した．実験の結果，2D動作であるにもかかわらず，複数のMoLMsで3D動作を入力した場合と同等の性能を維持できることが確認された．
%また，単眼実動画データセットを構築し，実動画における2D動作の3D動作に対する有用性を示した．また，実動画からの推定動作に含まれるノイズに対応するためにノイズ軽減アダプターを提案し，よりMoLMsの実応用性を向上させた．
%今後の研究方向として，まず実応用の観点からは，Sec.~\ref{}で述べたような実動画中のオクルージョンや激しい動作に起因する姿勢推定誤差への対処が重要である．また，モデル能力の拡張という観点からは，大規模なキャプション付き人体動作動画データを用いてMoLMsの言語モデルを追加学習することで，動作語彙の拡張やより詳細な動作理解の実現が期待される．

\bibliographystyle{splncs04}
\bibliography{main}

@String(IJCV  = {Int. J. Comput. Vis.})

@String(CVPR  = {IEEE Conf. Comput. Vis. Pattern Recog.})

@String(ICCV  = {Int. Conf. Comput. Vis.})

@String(ECCV  = {Eur. Conf. Comput. Vis.})

@String(NeurIPS = {Adv. Neural Inform. Process. Syst.})

@String(ICLR  = {Int. Conf. Learn. Represent.})

@String(CVPRW = {IEEE Conf. Comput. Vis. Pattern Recog. Worksh.})

@String(AAAI  = {AAAI})

@String(TOG   = {ACM Trans. Graph.})

@String(IJCV  = {IJCV})

@String(CVPR  = {CVPR})

@String(ICCV  = {ICCV})

@String(ECCV  = {ECCV})

@String(NeurIPS = {NeurIPS})

@String(ICLR  = {ICLR})

@String(CVPRW = {CVPRW})

@String(TOG   = {ACM TOG})

@inproceedings{mgpt,
  author       = {Biao Jiang and
                  Xin Chen and
                  Wen Liu and
                  Jingyi Yu and
                  Gang Yu and
                  Tao Chen},
  title        = {MotionGPT: Human Motion as a Foreign Language},
  booktitle    = {NeurIPS},
  year         = {2023},
}

@inproceedings{llmsaregoodaction,
  author       = {Haoxuan Qu and
                  Yujun Cai and
                  Jun Liu},
  title        = {LLMs are Good Action Recognizers},
  booktitle    = {CVPR},
  pages        = {18395--18406},
  year         = {2024},
}

@article{mgpt2,
  author       = {Yuan Wang and
                  Di Huang and
                  Yaqi Zhang and
                  Wanli Ouyang and
                  Jile Jiao and
                  Xuetao Feng and
                  Yan Zhou and
                  Pengfei Wan and
                  Shixiang Tang and
                  Dan Xu},
  title        = {MotionGPT-2: {A} General-Purpose Motion-Language Model for Motion
                  Generation and Understanding},
  journal      = {CoRR},
  volume       = {abs/2410.21747},
  year         = {2024},
}

@article{mgpt3,
  author       = {Bingfan Zhu and
                  Biao Jiang and
                  Sunyi Wang and
                  Shixiang Tang and
                  Tao Chen and
                  Linjie Luo and
                  Youyi Zheng and
                  Xin Chen},
  title        = {MotionGPT3: Human Motion as a Second Modality},
  journal      = {CoRR},
  volume       = {abs/2506.24086},
  year         = {2025},
}

@article{mlm_daily,
  author       = {Gabriele Civitarese and
                  Michele Fiori and
                  Priyankar Choudhary and
                  Claudio Bettini},
  title        = {Large Language Models Are Zero-Shot Recognizers for Activities of
                  Daily Living},
  journal      = {{ACM} Trans. Intell. Syst. Technol.},
  volume       = {16},
  number       = {4},
  pages        = {78:1--78:32},
  year         = {2025},
}

@inproceedings{mgpt4,
  author       = {Yaqi Zhang and
                  Di Huang and
                  Bin Liu and
                  Shixiang Tang and
                  Yan Lu and
                  Lu Chen and
                  Lei Bai and
                  Qi Chu and
                  Nenghai Yu and
                  Wanli Ouyang},
  title        = {MotionGPT: Finetuned LLMs Are General-Purpose Motion Generators},
  booktitle    = {AAAI},
  pages        = {7368--7376},
  year         = {2024},
}

@inproceedings{chatpose,
  author       = {Yao Feng and
                  Jing Lin and
                  Sai Kumar Dwivedi and
                  Yu Sun and
                  Priyanka Patel and
                  Michael J. Black},
  title        = {ChatPose: Chatting about 3D Human Pose},
  booktitle    = {CVPR},
  pages        = {2093--2103},
  year         = {2024},
}

@article{t2mgpt,
  author       = {Jianrong Zhang and
                  Yangsong Zhang and
                  Xiaodong Cun and
                  Shaoli Huang and
                  Yong Zhang and
                  Hongwei Zhao and
                  Hongtao Lu and
                  Xi Shen},
  title        = {{T2M-GPT:} Generating Human Motion from Textual Descriptions with
                  Discrete Representations},
  journal      = {CoRR},
  volume       = {abs/2301.06052},
  year         = {2023},
}

@article{t2mgpt_hifi,
  author       = {Congyi Wang},
  title        = {T2M-HiFiGPT: Generating High Quality Human Motion from Textual Descriptions
                  with Residual Discrete Representations},
  journal      = {CoRR},
  volume       = {abs/2312.10628},
  year         = {2023},
}

@inproceedings{TM2T,
  author       = {Chuan Guo and
                  Xinxin Zuo and
                  Sen Wang and
                  Li Cheng},
  title        = {{TM2T:} Stochastic and Tokenized Modeling for the Reciprocal Generation
                  of 3D Human Motions and Texts},
  booktitle    = {ECCV},
  pages        = {580--597},
  year         = {2022},
}

@inproceedings{humanml3d,
  author       = {Chuan Guo and
                  Shihao Zou and
                  Xinxin Zuo and
                  Sen Wang and
                  Wei Ji and
                  Xingyu Li and
                  Li Cheng},
  title        = {Generating Diverse and Natural 3D Human Motions from Text},
  booktitle    = {CVPR},
  pages        = {5142--5151},
  year         = {2022},
}

@inproceedings{aist,
  author       = {Ruilong Li and
                  Shan Yang and
                  David A. Ross and
                  Angjoo Kanazawa},
  title        = {{AI} Choreographer: Music Conditioned 3D Dance Generation with {AIST++}},
  booktitle    = {ICCV},
  pages        = {13381--13392},
  year         = {2021},
}

@article{kit,
  author       = {Matthias Plappert and
                  Christian Mandery and
                  Tamim Asfour},
  title        = {The {KIT} Motion-Language Dataset},
  journal      = {Big Data},
  volume       = {4},
  number       = {4},
  pages        = {236--252},
  year         = {2016},
}

@inproceedings{coco,
  author       = {Tsung{-}Yi Lin and
                  Michael Maire and
                  Serge J. Belongie and
                  James Hays and
                  Pietro Perona and
                  Deva Ramanan and
                  Piotr Doll{\'{a}}r and
                  C. Lawrence Zitnick},
  title        = {Microsoft {COCO:} Common Objects in Context},
  booktitle    = {ECCV},
  pages        = {740--755},
  year         = {2014},
}

@inproceedings{vq-vae,
  author       = {A{\"{a}}ron van den Oord and
                  Oriol Vinyals and
                  Koray Kavukcuoglu},
  title        = {Neural Discrete Representation Learning},
  booktitle    = {NeurIPS},
  pages        = {6306--6315},
  year         = {2017},
}

@inproceedings{mg-motionllm,
  author       = {Bizhu Wu and
                  Jinheng Xie and
                  Keming Shen and
                  Zhe Kong and
                  Jianfeng Ren and
                  Ruibin Bai and
                  Rong Qu and
                  Linlin Shen},
  title        = {MG-MotionLLM: {A} Unified Framework for Motion Comprehension and Generation
                  across Multiple Granularities},
  booktitle    = {CVPR},
  pages        = {27849--27858},
  year         = {2025},
}

@inproceedings{momask,
  author       = {Chuan Guo and
                  Yuxuan Mu and
                  Muhammad Gohar Javed and
                  Sen Wang and
                  Li Cheng},
  title        = {MoMask: Generative Masked Modeling of 3D Human Motions},
  booktitle    = {CVPR},
  pages        = {1900--1910},
  year         = {2024},
}

@inproceedings{m3gpt,
  author       = {Mingshuang Luo and
                  Ruibing Hou and
                  Zhuo Li and
                  Hong Chang and
                  Zimo Liu and
                  Yaowei Wang and
                  Shiguang Shan},
  title        = {M$^3$GPT: An Advanced Multimodal, Multitask
                  Framework for Motion Comprehension and Generation},
  booktitle    = {NeurIPS},
  year         = {2024},
}

@article{rapverse,
  author       = {Jiaben Chen and
                  Xin Yan and
                  Yihang Chen and
                  Siyuan Cen and
                  Qinwei Ma and
                  Haoyu Zhen and
                  Kaizhi Qian and
                  Lie Lu and
                  Chuang Gan},
  title        = {RapVerse: Coherent Vocals and Whole-Body Motions Generations from
                  Text},
  journal      = {ICCV},
  year         = {2025},
}

@article{socialgen,
  author       = {Heng Yu and
                  Juze Zhang and
                  Changan Chen and
                  Tiange Xiang and
                  Yusu Fang and
                  Juan Carlos Niebles and
                  Ehsan Adeli{-}Mosabbeb},
  title        = {SocialGen: Modeling Multi-Human Social Interaction with Language Models},
  journal      = {CoRR},
  volume       = {abs/2503.22906},
  year         = {2025},
}

@article{intergen,
  author       = {Han Liang and
                  Wenqian Zhang and
                  Wenxuan Li and
                  Jingyi Yu and
                  Lan Xu},
  title        = {InterGen: Diffusion-Based Multi-human Motion Generation Under Complex
                  Interactions},
  journal      = {IJCV},
  volume       = {132},
  number       = {9},
  pages        = {3463--3483},
  year         = {2024},
}

@inproceedings{TSMM,
  author       = {Mengyi Shan and
                  Lu Dong and
                  Yutao Han and
                  Yuan Yao and
                  Tao Liu and
                  Ifeoma Nwogu and
                  Guo{-}Jun Qi and
                  Mitch Hill},
  title        = {Towards Open Domain Text-Driven Synthesis of Multi-person Motions},
  booktitle    = {ECCV},
  pages        = {67--86},
  year         = {2024},
}

@article{3dpose_estim1,
  author       = {Muyu Li and
                  Henan Hu and
                  Jingjing Xiong and
                  Xudong Zhao and
                  Hong Yan},
  title        = {TSwinPose: Enhanced monocular 3D human pose estimation with JointFlow},
  journal      = {Expert Syst. Appl.},
  volume       = {249},
  pages        = {123545},
  year         = {2024},
}

@inproceedings{3dpose_estim2,
  author       = {Hongwei Zheng and
                  Han Li and
                  Wenrui Dai and
                  Ziyang Zheng and
                  Chenglin Li and
                  Junni Zou and
                  Hongkai Xiong},
  title        = {HiPART: Hierarchical Pose AutoRegressive Transformer for Occluded
                  3D Human Pose Estimation},
  booktitle    = {CVPR},
  pages        = {16807--16817},
  year         = {2025},
}

@inproceedings{3dpose_estim3,
  author       = {Jia Gong and
                  Lin Geng Foo and
                  Zhipeng Fan and
                  Qiuhong Ke and
                  Hossein Rahmani and
                  Jun Liu},
  title        = {DiffPose: Toward More Reliable 3D Pose Estimation},
  booktitle    = {CVPR},
  pages        = {13041--13051},
  year         = {2023},
}

@inproceedings{3dpose_estim4,
  author       = {Zhenhua Tang and
                  Zhaofan Qiu and
                  Yanbin Hao and
                  Richang Hong and
                  Ting Yao},
  title        = {3D Human Pose Estimation with Spatio-Temporal Criss-Cross Attention},
  booktitle    = {CVPR},
  pages        = {4790--4799},
  year         = {2023},
}

@inproceedings{3dpose_estim5,
  author       = {Ce Zheng and
                  Sijie Zhu and
                  Mat{\'{\i}}as Mendieta and
                  Taojiannan Yang and
                  Chen Chen and
                  Zhengming Ding},
  title        = {3D Human Pose Estimation with Spatial and Temporal Transformers},
  booktitle    = {ICCV},
  pages        = {11636--11645},
  year         = {2021},
}

@article{free3d,
  author       = {Sheng Liu and
                  Yuanzhi Liang and
                  Sidan Du},
  title        = {Free3D: 3D Human Motion Emerges from Single-View 2D Supervision},
  journal      = {CoRR},
  volume       = {abs/2511.11368},
  year         = {2025},
}

@inproceedings{v-vipe,
  author       = {Mara Levy and
                  Abhinav Shrivastava},
  title        = {{V-VIPE:} Variational View Invariant Pose Embedding},
  booktitle    = {CVPRW},
  pages        = {1633--1642},
  year         = {2024},
}

@inproceedings{unified2d3d,
  title={Unified 2D-3D Discrete Priors for Noise-Robust and Calibration-Free Multiview 3D Human Pose Estimation},
  author={Chen, Geng and Ren, Pengfei and Jian, Xufeng and Sun, Haifeng and Zhang, Menghao and Qi, Qi and Zhuang, Zirui and Wang, Jing and Liao, Jianxin and Wang, Jingyu},
  booktitle={NeurIPS},
  year         = {2025},
}

@article{t5,
  author       = {Colin Raffel and
                  Noam Shazeer and
                  Adam Roberts and
                  Katherine Lee and
                  Sharan Narang and
                  Michael Matena and
                  Yanqi Zhou and
                  Wei Li and
                  Peter J. Liu},
  title        = {Exploring the Limits of Transfer Learning with a Unified Text-to-Text
                  Transformer},
  journal      = {J. Mach. Learn. Res.},
  volume       = {21},
  pages        = {140:1--140:67},
  year         = {2020},
}

@article{Loper2015SMPL,
  title     = {SMPL: A Skinned Multi-Person Linear Model},
  author    = {Loper, Matthew and Mahmood, Naureen and Romero, Javier and Pons-Moll, Gerard and Black, Michael J.},
  journal   = {ACM Transactions on Graphics (TOG)},
  volume    = {34},
  number    = {6},
  year      = {2015}
}

@inproceedings{amass,
  author       = {Naureen Mahmood and
                  Nima Ghorbani and
                  Nikolaus F. Troje and
                  Gerard Pons{-}Moll and
                  Michael J. Black},
  title        = {{AMASS:} Archive of Motion Capture As Surface Shapes},
  booktitle    = {ICCV},
  pages        = {5441--5450},
  year         = {2019},
}

@inproceedings{finemotion,
  author       = {Bizhu Wu and
                  Jinheng Xie and
                  Meidan Ding and
                  Zhe Kong and
                  Jianfeng Ren and
                  Ruibin Bai and
                  Rong Qu and
                  Linlin Shen},
  title        = {FineMotion: {A} Dataset and Benchmark with Both Spatial and Temporal
                  Annotation for Fine-Grained Motion Generation and Editing},
  booktitle    = {ICCV},
  pages        = {13837--13846},
  year         = {2025},
}

@inproceedings{bleu,
  title={Bleu: a method for automatic evaluation of machine translation},
  author={Papineni, Kishore and Roukos, Salim and Ward, Todd and Zhu, Wei-Jing},
  booktitle={Proceedings of the 40th annual meeting of the Association for Computational Linguistics},
  pages={311--318},
  year={2002}
}

@inproceedings{rouge,
  title={Rouge: A package for automatic evaluation of summaries},
  author={Lin, Chin-Yew},
  booktitle={Text summarization branches out},
  pages={74--81},
  year={2004}
}

@inproceedings{cider,
  title={Cider: Consensus-based image description evaluation},
  author={Vedantam, Ramakrishna and Lawrence Zitnick, C and Parikh, Devi},
  booktitle={Proceedings of the IEEE conference on computer vision and pattern recognition},
  pages={4566--4575},
  year={2015}
}

@inproceedings{bertscore,
  author       = {Tianyi Zhang and
                  Varsha Kishore and
                  Felix Wu and
                  Kilian Q. Weinberger and
                  Yoav Artzi},
  title        = {BERTScore: Evaluating Text Generation with {BERT}},
  booktitle    = {ICLR},
  publisher    = {OpenReview.net},
  year         = {2020},
}

@inproceedings{
  vitpose,
  title={Vi{TP}ose: Simple Vision Transformer Baselines for Human Pose Estimation},
  author={Yufei Xu and Jing Zhang and Qiming Zhang and Dacheng Tao},
  booktitle={NeurIPS},
  year={2022},
}

@inproceedings{TRACE,
  author       = {Yu Sun and
                  Qian Bao and
                  Wu Liu and
                  Tao Mei and
                  Michael J. Black},
  title        = {{TRACE:} 5D Temporal Regression of Avatars with Dynamic Cameras in
                  3D Environments},
  booktitle    = {CVPR},
  pages        = {8856--8866},
  year         = {2023},
}

@inproceedings{motion-x,
  author       = {Jing Lin and
                  Ailing Zeng and
                  Shunlin Lu and
                  Yuanhao Cai and
                  Ruimao Zhang and
                  Haoqian Wang and
                  Lei Zhang},
  title        = {Motion-X: {A} Large-scale 3D Expressive Whole-body Human Motion Dataset},
  booktitle    = {NeurIPS},
  year         = {2023},
}

@misc{pyrender,
author = {Matthew Matl},
title = {Pyrender},
year = {2019},
publisher = {GitHub},
journal = {GitHub repository},
}

@inproceedings{WHAM,
  author       = {Soyong Shin and
                  Juyong Kim and
                  Eni Halilaj and
                  Michael J. Black},
  title        = {{WHAM:} Reconstructing World-Grounded Humans with Accurate 3D Motion},
  booktitle    = {CVPR},
  year         = {2024},
}

@article{distill2d3d,
  author       = {Xiangyue Zhang and
                  Yifan Jia and
                  Jiaxu Zhang and
                  Yijie Yang and
                  Zhigang Tu},
  title        = {Robust 2D Skeleton Action Recognition via Decoupling and Distilling
                  3D Latent Features},
  journal      = {{IEEE} Trans. Circuits Syst. Video Technol.},
  volume       = {35},
  number       = {10},
  pages        = {10410--10422},
  year         = {2025},
}

\clearpage
\clearpage

\setcounter{section}{0}
\setcounter{subsection}{0}
\setcounter{figure}{0}
\setcounter{table}{0}
\setcounter{equation}{0}

\renewcommand{\thesection}{\Alph{section}}
\renewcommand{\thefigure}{S\arabic{figure}}
\renewcommand{\thetable}{S\arabic{table}}
\renewcommand{\theequation}{S\arabic{equation}}

\section*{Supplementary Material}

This supplementary material provides training and implementation details, as well as additional experimental results, that could not be included in the main paper due to space limitations.
Specifically, Sec.~\ref{sec:imple} presents additional implementation details of the 2D motion encoder and the implementation details of each baseline model. Section.~\ref{sec:qualitative_2d} presents qualitative results for motion captioning and motion-to-detailed text. Section.~\ref{sec:supp_analysis} provides additional analyses of the 2D motion encoder through multi-view token consistency and 3D motion reconstruction results.  In addition, Sec.~\ref{sec:real_video_dataset} describes the details of the monocular real-world video motion dataset, and Sec.~\ref{sec:training_adapter} presents the training and implementation details of the real-video adapter, along with an ablation study on adapter architectures and qualitative results. %Finally,  details of the supplementary code package are summarized in Sec.~\ref{sec:code}.
 
%この補足資料では，紙面の都合により本論文に含めることができなかった学習および実装の詳細，実験結果に関する情報を示す．

%具体的には，Sec.~\ref{sec:imple}では2D motion encoderの追加の実装の詳細や，各ベースラインモデルの実装の詳細を示し，Sec.~\ref{sec:qualitative_2d}においてmotion captioning, motion-to-detailed textの定性結果を示す．
%さらに，Sec.~\ref{sec:real_video_dataset}では，monocular real-world video motion datasetの詳細，Sec.~\ref{training_adapter}ではreal-video adapterの学習，実装の詳細，定性結果を示す．最後に，Sec.~\ref{sec:code}ではコードパッケージと再現性に関する詳細をまとめる．

%%%%%%%%%%%%%%%%%%%%%%%%%%%%%%
\section{Additional Implementation Details.}
\label{sec:imple}

\subsection{2D motion encoder $E_\mathrm{2D}$}
\label{subsec:2dmotionencoder}

The 2D encoder adopts the same architecture as the VQ-VAE encoder of MotionGPT\cite{mgpt}.

\noindent
\textbf{Training time.} Training takes around 22 hours (3000 epochs). During training, we apply view randomization by sampling yaw angles from the full 360-degree range and pitch angles from 0 to 60 degrees. 
\noindent
\textbf{Parameters.} The 2D motion encoder $E_\mathrm{2D}$ has 9.6M parameters.

\subsection{2D Motion Normalization}
The 68-dimensional feature $\bm{X}^{\text{2D}}$ is normalized in two stages.

\textit{(i) Per-clip scale normalization.}
Since orthographic projection leaves the absolute scale of the subject
undetermined, we first remove clip-level scale and translation.
Let $\bm{r}_t=\frac{1}{2}(\bm{j}^{\text{left\_hip}}_t+\bm{j}^{\text{right\_hip}}_t)$ be the root and $\tilde{\bm{j}}_{t,i}=\bm{j}_{t,i}-\bm{r}_t$ the root-relative
joint positions. We define a single scalar scale per clip as
\begin{equation}
s=\max\Big(P_{99}\big(|\tilde{x}|\big),\;P_{99}\big(|\tilde{y}|\big),\;\epsilon\Big),
\end{equation}
where $P_{99}(\cdot)$ denotes the $99$th percentile taken over all $T$ frames
and all $13$ joints of the clip, and $\epsilon=10^{-8}$ guards against
degenerate clips. Using a percentile rather than the maximum makes $s$ robust
to outlier joints, which matters for the noisy 2D keypoints obtained from a
real-video pose estimator. All translation-dependent quantities
(root $y$ position, joint positions, root velocity, joint velocity) are divided
by $s$; joint orientations are computed as $\arctan2$ of the normalized
root-relative coordinates and are scale-invariant by construction.
We further subtract the first-frame value from the root $y$ position so that
the representation does not depend on the absolute vertical placement of the
subject in the image. As a result, $\bm{X}^{\text{2D}}$ is invariant to the
subject's apparent size, camera distance, and in-image position.

\textit{(ii) Dataset-level z-normalization.}
We then apply a per-dimension z\hyp{}normalization
$\bm{X}^{\text{2D}}\!\leftarrow\!(\bm{X}^{\text{2D}}-\bm{\mu})/\bm{\sigma}$,
where $\bm{\mu},\bm{\sigma}\in\mathbb{R}^{68}$ are computed once over the
training split of HumanML3D and reused unchanged at evaluation time, so no
test-set statistics leak into the model.

\subsection{Baseline Models}
\label{sec:baseline}
For all baseline models, the results are obtained using the official implementations. However, the checkpoints used differ across models and settings (e.g., 3D Input, 2D Scratch and Ours).
%すべてのベーラインモデルにおいて，結果は公式実装を用いて得られています．しかし，使用したチェックポイントはモデル間，メソッド間（e.g., 3D Input, 2D from scrach)で差異があります．

\textbf{3D Inputs.}
While MotionGPT and MG-MotionLLM~\cite{mg-motionllm} use the same VQ-VAE architecture, TM2T~\cite{TM2T} adopts a different VQ-VAE architecture. Therefore, in this work, we modify TM2T to use the same VQ-VAE architecture as the other models and retrain the entire model. This allows us to focus on the effect of changing the input modality from 3D motions to 2D motions, rather than differences in model-specific VQ-VAE designs. For MotionGPT and MG-MotionLLM, we use the officially released checkpoints.
%MotionGPTおよびMG-MotionLLMでは同一のVQ-VAEアーキテクチャが用いられている一方，TM2Tでは異なるVQ-VAEアーキテクチャが採用されている．そのため，本研究ではTM2Tについても他のモデルと同一のVQ-VAEアーキテクチャを用いるように変更し，モデル全体を再学習した．これにより，モデル固有のVQ-VAE設計の違いではなく，入力モダリティを3D動作から2D動作へ変更したことによる影響に焦点を当てて評価できる．MotionGPTとMG-MotionLLMでは公式に配布されているチェックポイントを用いています

\textbf{2D Scratch.}
For MG-MotionLLM and TM2T, we retrain the entire model following the official implementations after replacing the input modality with 2D motions. For TM2T, as described above, we modify the VQ-VAE architecture to match that used in the other models. In contrast, for MotionGPT, we do not report the 2D-from-scratch result because we could not stably reproduce the training performance using the official implementation. We note that the reproducibility issue regarding MotionGPT retraining has also been reported in the official repository issues and remains unresolved at the time of writing.
%MG-MotionLLMおよびTM2Tについては，公式実装に従い，入力モダリティを2D動作へ変更した上でモデル全体を再学習した．ただし，TM2Tについては前述の理由により，VQ-VAEアーキテクチャを他のモデルと統一している．一方，MotionGPTについては，公式実装を用いた再学習において安定した性能を再現できなかったため，2D動作からのスクラッチ学習結果は報告しない．なお，MotionGPTの学習に関する再現性の問題は公式リポジトリのissueでも報告されており，現時点で解決されていない．

\textbf{Ours.}
As in the 3D Input setting, we use the officially released checkpoints for MotionGPT and MG-MotionLLM, and use the checkpoint obtained by retraining TM2T with the modified VQ-VAE architecture.
%3D Inputsの場合と同様に，MotionGPTとMG-MotionLLMでは公式に配布されているチェックポイントを用い,TM2TはVQ-VAEのアーキテクチャを変更して再学習した場合に得られたチェックポイントを用いています．

\subsection{Inference Details}
\label{subsec:inference_2d}

\noindent
\textbf{Task prompt for motion understanding tasks.}
MG-MotionLLM and MotionGPT feed the discrete token sequence generated from the motion into the language model together with a task prompt. Specifically, the task prompt \texttt{Generate text:} is used for motion captioning, while \texttt{Generate the motion script:} is used for motion-to-detailed text.
%MotionGPTと，MG-MotionLLMでは動作から生成された離散トークン列をタスクプロンプトとともに言語モデルへ入力します．motion captioningにおいては'Generate text:'，motion-to-detailed textにおいては'Generate the motion script:'というタスクプロントが用いられます．
%%%%%%%%%%%%%%%%%%%%%%%%%%%%%%
\begin{figure}[t]
    \centering
    \includegraphics[width=\linewidth]{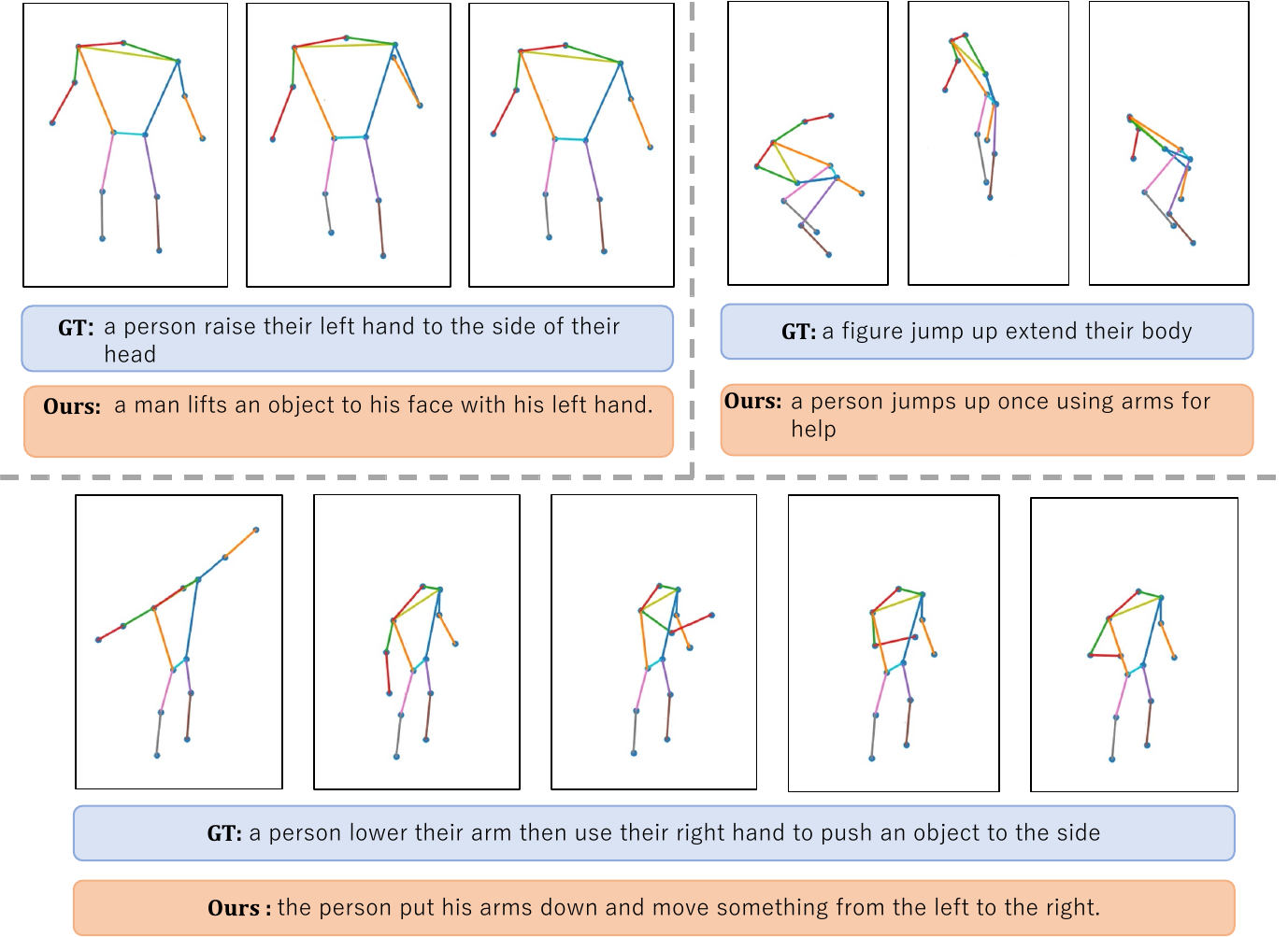}
    \vspace*{-5mm}
    \caption{Qualitative results of motion captioning on HumanML3D test set.}
    \label{fig:m2t_qual}
    \vspace*{-2mm}
\end{figure}

\section{Qualitative Results of motion-understanding tasks}
\label{sec:qualitative_2d}

\subsection{Results of motion captioning}
\label{subsec:qualo_2d_encoder_m2t}

The main paper reports quantitative motion captioning and motion-to-detailed text results in Sec.~4.2.
Here, we provide additional qualitative examples to complement those benchmark results.
%本文のSec.~4.2では，motion captioningおよびmotion-to-detailed textに関する定量的評価結果を示した。ここでは，それらのベンチマーク結果を補足するため，追加の定性的な生成例を示す。

Figure.~\ref{fig:m2t_qual} shows qualitative results of the proposed method on the motion captioning task using MotionGPT as the baseline model on HumanML3D~\cite{humanml3d}. The upper-left and upper-right examples in Fig.~\ref{fig:m2t_qual} correspond to relatively simple motions. In both examples, the generated captions are semantically consistent with the ground-truth (GT) captions. The bottom example in Fig.~\ref{fig:m2t_qual} shows a more complex motion composed of two primitive motions. In this case as well, the generated caption is semantically aligned with the GT caption. These qualitative results are consistent with the quantitative results reported in the main paper and support the conclusion that the proposed method largely preserves motion captioning performance even when using 2D motions as input.

%図~\ref{fig:m2t_qual}にHumanML3Dにおける，MotionGPTをベースラインモデルとして用いた場合の提案手法のmotion captioningタスクの定性結果を示す．図~\ref{fig:m2t_qual}の左上および右上の例は，比較的単純な動作に対応している。いずれの例においても，正解キャプションと生成したキャプションが意味的に整合していることが確認できる。
%また，図~\ref{fig:m2t_qual}の下段の例は，2つの基本動作から構成されるより複雑な動作を示している。この場合においても，正解キャプションと生成キャプションは意味的に一致していることが確認できる。
%これらの定性的結果は，本文で示した定量評価結果と整合しており，提案手法によって2D動作を入力とした場合でもmotion captioning性能を概ね維持できることを裏付けている．

\subsection{Results of motion-to-detailed text}
\label{subsec:qualo_2d_encoder_m2dt}

\begin{figure}[t]
    \centering
    \includegraphics[width=0.7\linewidth]{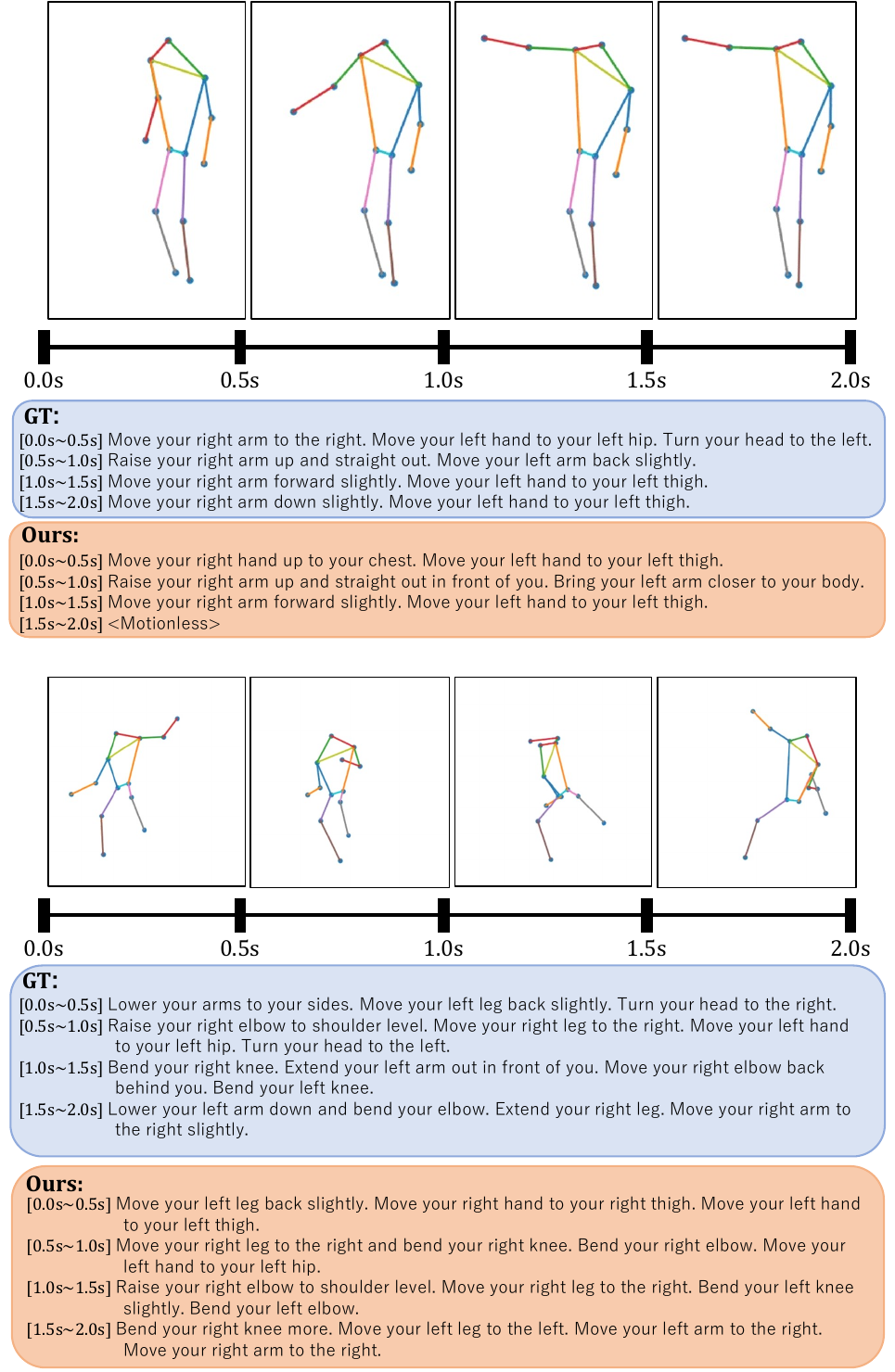}
    \vspace{-3mm}
    \caption{Qualitative results of motion-to-detailed text on FineMotion test set.}
    \label{fig:m2dt_qual}
    \vspace*{-2mm}
\end{figure}

Figure.~\ref{fig:m2dt_qual} shows qualitative results of the proposed method on the motion-to-detailed text task using MG-MotionLLM as the baseline model on FineMotion. In both examples, the captions generated by the proposed method are semantically consistent with the GT captions, which is consistent with the quantitative results reported in the main paper. These results demonstrate that the proposed method can effectively preserve not only coarse motion information, as in motion captioning, but also fine-grained snippet-level motion information.
%図~\ref{fig:m2dt_qual}に，FineMotionにおける，MG-MotionLLMをベースラインモデルとして用いた場合の提案手法のmotion-to-detailed text多数の定性結果を示す．両方の例でGTのキャプションと提案手法の生成キャプションが意味的に一致していることが確認でき，本文の定量評価結果と整合していることがわかる．提案手法によって，motion-captioningのようなcoarseな動作情報だけでなく，スニペットレベルの細粒度な動作情報も有効に保持できることを示している．

\section{Additional analysis of 2D motion encoder}
\label{sec:supp_analysis}

\subsection{3D reconstruction ability}
\label{Sec:reconstruction}

In Sec. 4.3 of the main paper, we showed that the discrete token sequences obtained from 2D motions tend to select neighboring tokens of those obtained from 3D motions, resulting in only limited semantic degradation. To further support this observation, this section compares the reconstructed 3D motions obtained by feeding the discrete token sequences derived from 3D motions and 2D motions into the VQ-VAE decoder. 
To evaluate 3D reconstruction quality, we report joint-position reconstruction errors with respect to the GT 3D motion $\bm{X}^{\mathrm{3D}}$, including Mean Per Joint Position Error (MPJPE) and Procrustes-Aligned MPJPE (PAMPJPE). We also report distribution- and motion-quality metrics, including FID, Diversity (Div), and ACCEL. The experiments are conducted using the VQ-VAE of MotionGPT and its corresponding 2D motion encoder. 
%本文のSec,~\ref{}では2D動作から得られた離散トークン列は，3D動作から得られた離散トークン列の近傍トークンを取っており，意味的な破壊が小さいことを述べた．本節ではその補強のために，3D動作から得られた離散トークン列と2D動作から得られた離散トークン列をVQ-VAEのデコーダに入力し，再構成された3D動作を比較する．

As shown in Tab.~\ref{tab:reconstruction}, the MPJPE of the 2D input is substantially larger than that of the 3D input. MPJPE measures the Euclidean distance between the corresponding joints of the predicted and GT 3D motions in the world coordinate system. This result indicates that the discrete motion tokens obtained from 2D motions do not necessarily preserve 3D geometric information, such as absolute positions in the global coordinate system.
In contrast, the gap in PAMPJPE is small, indicating that the reconstructed pose structure is largely preserved. Since PAMPJPE computes MPJPE after rigidly aligning the predicted and GT skeletons, it is insensitive to global position and orientation. This result demonstrates that the proposed 2D motion encoder successfully reconstructs the underlying pose structure.
Moreover, the distribution-based metrics, including FID, Diversity, and ACCEL, remain comparable between the 2D and 3D inputs. Motion semantics are generally considered to be more strongly associated with pose structure than with absolute positions in the global coordinate system. Therefore, these results support our conclusion that the discrete motion tokens obtained from 2D motions preserve semantic information with only minimal degradation.
%Tab.\~\ref\{}に示すように，2D Inputでは3D Inputと比較してMPJPEが大きく増加している。MPJPEは，予測された3D動作と正解3D動作における対応関節間のユークリッド距離を，ワールド座標系で測定する指標である。したがって，この結果は，2D動作から得られた離散トークン列では，グローバル座標系における位置などの3D幾何情報が必ずしも正確には復元されないことを示している。 一方で，PAMPJPEの差は小さく，再構成された動作の姿勢構造は概ね保持されていることが分かる。PAMPJPEは，予測スケルトンと正解スケルトンを最適に位置合わせした後にMPJPEを計算するため，グローバルな位置や姿勢の影響を取り除いた評価指標である。この結果は，2D motion encoderが姿勢構造を十分に再構成できることを示している。 また，FID，Diversity，ACCELなどの分布ベースの評価指標では，2D Inputは3D Inputと同程度の性能を維持している。動作の意味情報は，グローバル座標系における絶対位置よりも，姿勢構造により強く内包されると考えられる。したがって，以上の結果は，2D動作から得られた離散トークン列における意味的な破壊が小さいという結論を裏付けている。 
\begin{table}[t]
\centering
\caption{Comparison of 3D reconstruction ability using MotionGPT's VQ-VAE. The best and second-best results are highlighted in bold and underlined,
respectively.}
\label{tab:reconstruction}

\small
\begin{tabular}{l ccccc}
\toprule
 Method
& MPJPE$\downarrow$ & PAMPJPE$\downarrow$ & FID$\downarrow$ & Div$\uparrow$ & ACCEL$\downarrow$ \\
\midrule
 3DInput   &\textbf{57.5} &\textbf{41.4} &\textbf{0.083} &\underline{9.32} &\textbf{8.19} \\
 Ours (2D Input) &\underline{74.0} &\underline{43.1} &\underline{0.110} &\textbf{9.84} &\underline{8.32} \\
\bottomrule
\end{tabular}
\end{table}

\subsection{Viewpoint Sensitivity via Multi-view Token Consistency}
\label{subsec:viewpoint}

\begin{figure}[t]
    \centering
    \includegraphics[width=\linewidth]{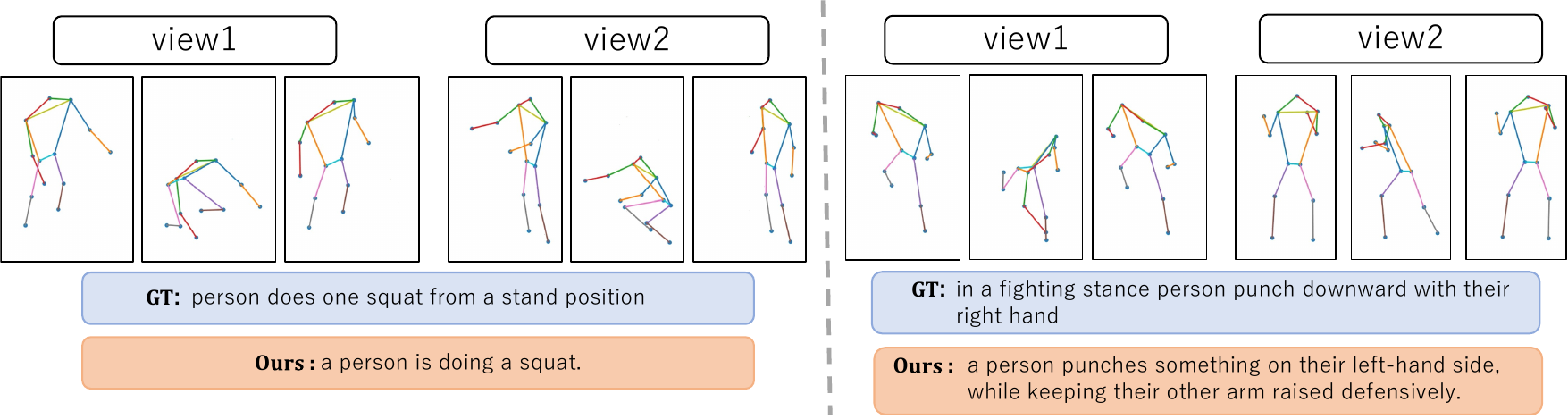}
    \vspace*{-5mm}
    \caption{Qualitative results of viewpoint robustness. The same 3D motion is observed from two viewpoints, while the generated captions remain consistent.}
    \label{fig:view_robust}
\end{figure}

In Sec.~3.2 of the main paper, view randomization is introduced to encourage the 2D motion encoder to learn representations that are robust to viewpoint changes. To verify this effect, this section analyzes how consistently discrete token sequences are obtained when multiple 2D motions generated from different viewpoints of the same 3D motion are used as inputs.
Specifically, given a 3D motion $\bm{X}^{\mathrm{3D}}$, we randomly sample $N$ different viewpoints and generate the corresponding 2D motions using the same procedure as in Sec.~3.2 of the main paper. We then feed each 2D motion into the 2D motion encoder and quantize the resulting latent features using the VQ-VAE codebook, obtaining a discrete $\bm{C}^{\mathrm{2D}}_i = [c^{\mathrm{2D}}_{i,1}, \ldots, c^{\mathrm{2D}}_{i,S}]$ for each viewpoint $v_i$. 

To evaluate multi-view token consistency, we examine whether the tokens obtained from all $N$ viewpoints are identical at each temporal position. Specifically, for each temporal index $s$, we regard the token as consistent if $c^{\mathrm{2D}}_{1,s} = c^{\mathrm{2D}}_{2,s} = \cdots = c^{\mathrm{2D}}_{N,s}$. The consistency score is computed as the proportion of temporal positions satisfying this condition, averaged over all test motions. Under random token assignment from a codebook of size $K$, the expected consistency is $(1/K)^{N-1}$. 
%本文のSec.~3.2では，view randomizationを導入することで，2D motion encoderが視点変化に対して頑健な表現を獲得することを意図している．本節では，その効果を検証するため，同一の3D動作から異なる視点で生成した複数の2D動作を入力した際に，得られる離散トークン列がどの程度一致するかを分析する．
%具体的には，3D動作 $\bm{X}^{\mathrm{3D}}$ に対して，$k$ 個の異なる視点をランダムにサンプリングし，本文のSec.~3.2と同様の手順で，各視点に対応する2D動作 ${\bm{X}^{\mathrm{2D}, v_i}}{i=1}^{k}$ を生成する．次に，各2D動作を2D motion encoderに入力し，得られた潜在特徴をVQ-VAEのコードブックで量子化することで，視点ごとの離散トークン列 ${\bm{C}^{\mathrm{2D}, v_i}}_{i=1}^{k}$ を得る．最後に，全ての視点ペアについて，本文のSec.~4.3と同様にTop-$k$ token agreementを計算し，異なる視点間で生成されるトークン列の一貫性を評価する．

The experiments were conducted using the VQ-VAE of MotionGPT and its corresponding 2D motion encoder. The agreement rates of the discrete token sequences were 63.6\% for $N=2$, 50.9\% for $N=3$, and 39.7\% for $N=5$. When tokens are randomly selected from the codebook, the expected agreement rate is $1/K \simeq 0.20\%$ when $N=2$; for $N=3$, this probability is squared, and for $N=5$, it is squared again. These results indicate that relatively consistent discrete token sequences can be obtained even when the viewpoint is changed. In addition, Fig.~\ref{fig:view_robust} provides complementary qualitative examples showing that this robustness is also reflected at the caption level. As shown in Fig.~\ref{fig:view_robust}, the predicted captions remain consistent across different viewpoints.
%実験ではMotionGPTのVQ-VAEとそれに対応する2D motion encoderを用いて実施された．離散トークン列の一致率はN=2の場合は63.6%，N=3の場合は50.9%, N=5の場合は39.7%であった．ランダムにトークンを選択した場合の期待一致率はN=2の時に，$1/K \simeq 0.20\%$，Top-3で $3/K \simeq 0.59\%$，N=3のときはその2乗，N=5の時はさらにその2乗である．この結果から，視点を変化させた場合においても，ある程度一貫した離散トークン列を得ることができることがわかる．また，Fig.~{}には，この頑健性がcaptionレベルにも反映されていることを示す補足的な定性例を示す．Fig.~\ref{fig}に示すように，予測されたcaptionは異なる視点間で一貫している．

%%%%%%%%%%%%%%%%%%%%%%%%%%%%%%
\section{Details of the Monocular Real-World Video Motion Dataset }
\label{sec:real_video_dataset}

%\subsection{Clip Sampling from HumanML3D}
We randomly extract motion clips from HumanML3D test set. %The list of sampled motion-clip filenames is included as part of the supplementary material; see \verb|sampled_motion_clip_for_realvideos.txt|.

Motion-X~\cite{motion-x}, a motion-language dataset, includes a subset of samples in which videos are paired with estimated 2D and 3D motions. Therefore, motion understanding tasks on real-world videos could also be performed using Motion-X. However, the captions annotated in Motion-X differ substantially in style from those in HumanML3D; for example, they may include emotional expressions. This introduces a domain gap, making it difficult to conduct a fair evaluation. For this reason, we do not use Motion-X for evaluation.

To assess the validity of the collected dataset, we conduct a human verification study. Three annotators are asked to judge whether each collected video is semantically consistent with its corresponding HumanML3D caption using a binary yes/no label. On average, 86.4\% of the video-caption pairs are judged as consistent. This result indicates that, although the videos are collected by asking participants to imitate HumanML3D motions, the original captions remain largely valid for evaluating motion understanding on the collected real-world videos. The remaining inconsistent cases may be due to imitation errors during data collection or mismatches between the original HumanML3D motions and their captions.
%モーション言語データセットであるMotion-Xには一部，動画とそこから推定された2D, 3D動作がペアになっているものがあるので，Motion-Xでも実環境動画における動作理解タスクは実行可能であるが，Motion-XにアノテーションされてキャプションはHumanML3Dと比較して，長かったり，感情表現を含むなどキャプションスタイルが大きくなり，ドメインギャップの問題が発生して正しく評価することができないため評価には用いなかった．

All performers participated with informed consent.
%We plan to publicly release the collected videos and the corresponding pose-estimation results. Typical examples from our dataset, including videos and the corresponding pose-estimation results, are included as part of the supplementary material; see \allowbreak{\verb|Monocular_Real_Videos_Dataset|.}

%%%%%%%%%%%%%%%%%%%%%%%%%%%%%%
\section{Additional details of real-video adapter}
\label{sec:training_adapter}

This section provides the details of the real-video adapter $A_{real}$, which are omitted from the main paper due to space limitations.

\subsection{Datasets for Training}
\label{subsec:adapter_datasets}
\noindent\textbf{Train/Val split.}
Since HumanML3D is built from AMASS~\cite{amass}, we sample AMASS motions according to the HumanML3D splits. Specifically, we use 3,262 motions from the HumanML3D train set for training and 813 motions from the HumanML3D validation set for validation. For each motion, we render \textbf{10} viewpoints.
The checkpoint that achieves the best performance on the validation set is used for evaluation on the monocular real-world video motion dataset.

\subsection{Implementation Details}
\label{subsec:adapter_impli}

We implement the adapter, $A_{real}$, as a lightweight residual MLP operating on the per-frame input features. Specifically, $A_\text{real}$ consists of a LayerNorm, a two-layer MLP with hidden dimension 512, GELU activation, and dropout $p{=}0.1$. The estimated 2D motion $\bm{X}_{\mathrm{PE}}^{\mathrm{2D}}$ and the per-joint confidence scores $\bm{Conf}$ are concatenated and fed into the adapter $A_{real}$.

%アダプタA_realは，各フレームの入力特徴に対して動作する軽量な残差MLPとして実装する。具体的には，A_realは，Layer Normalization，隠れ層次元512の2層MLP，GELU活性化関数，およびドロップアウト率 p=0.1 のドロップアウト層から構成される。estimated 2D motion X^PE_2Dと各関節の確信度ConfはコンキャットされてアダプターA_realに入力されます．

$A_{real}$ is trained using AdamW with learning rate = $1 \times 10^{-3}$, $\beta=(0.9,0.99)$, batch size = $64$, and weight decay = $1 \times 10^{-4}$, for 150k iterations, without a learning-rate scheduler.

%\subsection{Training Details}
%label{subsec:training_adapter}

\noindent
\textbf{GPU.} We train the adapter, $A_{real}$, on a single NVIDIA A100 40GB GPU.

\noindent
\textbf{Training time.} Training takes 1 hour (150k iterations).

\noindent
\textbf{Parameters.} The adapter, $A_{real}$, has 0.3M parameters.

\subsection{Real-video adapter for estimated 3D motions}
The adapter $A_{\mathrm{3D}}$ for estimated 3D motions is trained in the same manner as the adapter $A_{\mathrm{real}}$ for estimated 2D motions. Specifically, following the procedure described in Sec.~5 of the main paper, we generate pseudo-real training pairs using TRACE and WHAM, respectively, instead of ViTPose, and train the adapters with the same training objective. Note that, unlike $A_{\mathrm{real}}$, $A_{\mathrm{3D}}$ takes 3D motion features as input; therefore, its input dimensionality is 263 and no confidence input is used.
%推定された3DモーションのためのアダプターA3Dは推定された2Dモーション用のアダプターA_realと同様の方法で訓練されます．具体的には本文のSec. 5で紹介した方法と同様にTRACE, WHAMをViTPoseの代わりに用いてPseudo-real training pairsをそれぞれ生成し，同じTraining Objectiveを用いて訓練されます．ただし，A_realとは異なり，3Dモーションを用いるため，入力次元が263次元になり，confidence inputはないことに注意してください．

\subsection{Qualitative Evaluation}
\label{subsec:add_quali_real}

\begin{figure}[t]
    \centering
    \includegraphics[width=\linewidth]{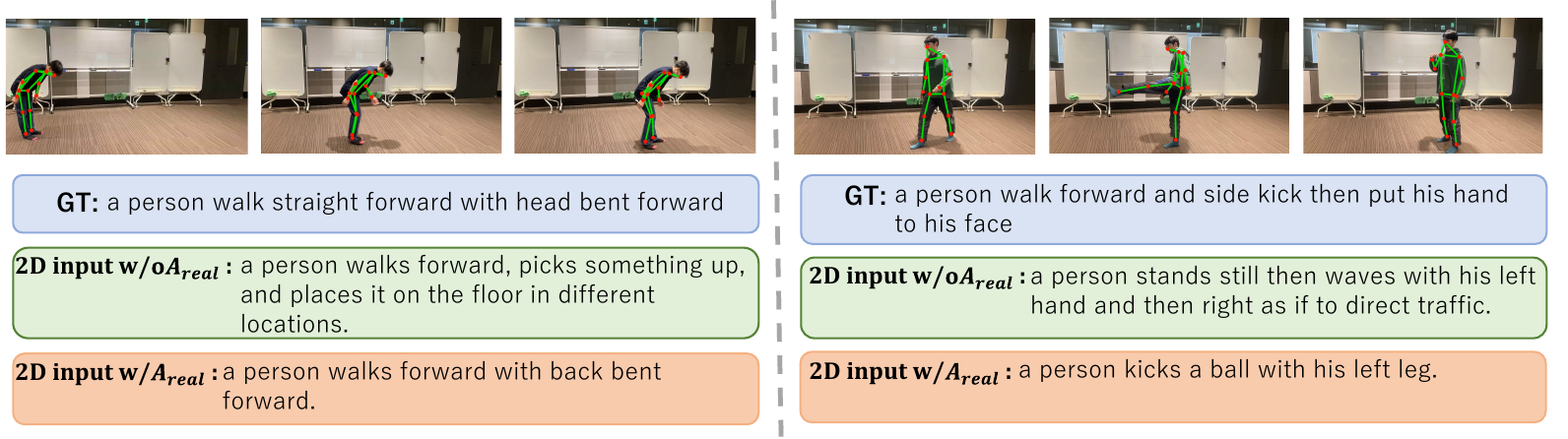}
    \vspace*{-5mm}
    \caption{Qualitative comparison with and without the adapter on monocular real videos.}
    \label{fig:adapter_quali}
\end{figure}

Figure.~\ref{fig:adapter_quali} shows qualitative results on the monocular real-world video motion dataset using MotionGPT as the baseline model. Here, we compare the generated captions with and without the adapter $A_{real}$.
In all examples, introducing the adapter makes the generated captions semantically closer to the GT captions. In contrast, without the adapter, the model sometimes fails to recognize the motion correctly even for relatively simple motions or scenes with little occlusion, generating captions that deviate substantially from the GT captions.

These results indicate that the adapter effectively mitigates the domain shift caused by pose-estimation noise and improves the reliability of the motion tokens fed into the language model of MotionGPT.
%Fig.~\ref{fig:adapter_quali}にMotionGPTをベースラインとして用いた場合のMonocular real-world video motion datasetにおける定性結果を示す．ここでは，アダプタを導入しない場合と導入した場合のキャプション生成結果を比較する．
%いずれの例においても，アダプタを導入することで，生成キャプションは正解（GT）キャプションにより近い意味内容となっていることが確認できる。
%一方，アダプタを用いない場合には，比較的単純な動作やオクルージョンの少ない場面であっても認識に失敗し，正解キャプションから大きく逸脱したキャプションを生成する場合が見られる。
%これらの結果は，アダプタA_realが姿勢推定ノイズによって生じるドメインシフトを効果的に軽減し，MotionGPTの言語モデルへ入力される動作トークンの信頼性を向上させていることを示している。

\subsection{Ablation Study on Adapter Architecture}
\label{subsec:ablation_adapter}
To investigate the effect of the design of the adapter $A_{real}$ on performance, we compare multiple adapter architectures in this section. Using MotionGPT as the baseline model, we evaluate a Linear adapter and a Temporal Conv adapter in addition to the MLP-based adapter used in the main paper. All adapters include residual connections and are trained and evaluated under the same settings as in Sec.~5 of the main paper.
%アダプタ \$A\_{\textbackslash{}mathrm\{real\}}\$ の設計が性能に与える影響を調べるため，本節では複数のアダプタアーキテクチャを比較する．MotionGPTをベースラインモデルとし，本文で用いたMLPベースのアダプタに加えて，Linear adapterおよびTemporal Conv adapterを評価する．全てのアダプタは残差接続を持ち，本文のSec.\~5と同一の学習・評価設定を用いる． 

\begin{table}[t]
\centering
\caption{Comparison of adapter architectures using MotionGPT.}
\label{tab:ablation_adapter}
\small
\begin{tabular}{l ccccc}
\toprule
 Architecture
& BLEU-1$\uparrow$ & BLEU-4$\uparrow$ & ROUGE-L$\uparrow$ & CIDEr$\uparrow$ & BERTScore$\uparrow$ \\
\midrule
 Linear   &\underline{0.413} &\textbf{0.057} &0.362 &\underline{0.105} &0.294 \\
 Temporal Conv &0.399 &\underline{0.053} &\textbf{0.378} &\textbf{0.110} &\underline{0.299} \\
 Ours (MLP)&\textbf{0.435} &0.045 &\underline{0.367} &\underline{0.105} &\textbf{0.306} \\
\bottomrule
\end{tabular}
\end{table} 

As shown in Tab.~\ref{tab:ablation_adapter}, all adapter architectures achieve comparable performance, indicating that the effectiveness of the real-video adapter does not strongly depend on a specific architectural design. Among them, the MLP adapter achieves the best BLEU-1 and BERTScore while remaining competitive on ROUGE-L and CIDEr. Therefore, we use the MLP adapter as the default architecture.

\end{document}